\documentclass{article}

\usepackage{microtype}
\usepackage{graphicx}
\usepackage{subcaption}
\usepackage{booktabs} 
\usepackage{multirow} 
\usepackage{array}

\usepackage{hyperref}

\usepackage[preprint]{icml2026}

\usepackage{amsmath}
\usepackage{amssymb}
\usepackage{mathtools}
\usepackage{amsthm}
\usepackage{booktabs}
\usepackage[table]{xcolor}
\usepackage{graphicx}
\usepackage{subcaption}
\usepackage{algorithm}

\usepackage[capitalize,noabbrev]{cleveref}

\theoremstyle{plain}

\theoremstyle{definition}

\theoremstyle{remark}

\usepackage{xcolor}
\definecolor{logored}{HTML}{DA3A3C}
\definecolor{logogreen}{HTML}{09A06D}

\usepackage[textsize=tiny]{todonotes}

\icmltitlerunning{Mind-Brush: Integrating Agentic Cognitive Search and Reasoning into Image Generation}

\begin{document}

\twocolumn[
  \icmltitle{Mind-Brush: Integrating Agentic Cognitive Search and  \\ Reasoning into Image Generation}



  \icmlsetsymbol{equal}{*}
  \icmlsetsymbol{corrs}{†}

\newcounter{@affilsysu}\setcounter{@affilsysu}{1}
\newcounter{@affilthu}\setcounter{@affilthu}{2}
\newcounter{@affilpjlab}\setcounter{@affilpjlab}{3}
\newcounter{@affilcuhk}\setcounter{@affilcuhk}{4}
\setcounter{@affiliationcounter}{4}
\makeatother

  \begin{icmlauthorlist}
    \icmlauthor{Jun He}{equal,sysu}
    \icmlauthor{Junyan Ye}{equal,sysu,pjlab}
    \icmlauthor{Zilong Huang}{sysu}
    \icmlauthor{Dongzhi Jiang}{cuhk}
    \icmlauthor{Chenjue Zhang}{thu} \\
    \icmlauthor{Leqi Zhu}{pjlab}
    \icmlauthor{Renrui Zhang}{cuhk}
    \icmlauthor{Xiang Zhang}{sysu}
    \icmlauthor{Weijia Li}{sysu,thu,corrs}
  \end{icmlauthorlist}

  \icmlaffiliation{sysu}{Sun Yat-sen University}
  \icmlaffiliation{thu}{Tsinghua Shenzhen International Graduate School, Tsinghua University}
  \icmlaffiliation{pjlab}{Shanghai Artificial Intelligence Laboratory}
  \icmlaffiliation{cuhk}{MMLab, The Chinese University of Hong Kong}

  \icmlcorrespondingauthor{Weijia Li}{liweij29@mail.sysu.edu.cn}

  \icmlkeywords{Machine Learning, ICML}

  \parbox{\textwidth}{
        \centering
        \begin{tabular}{ll}
        \raisebox{-0.15em}{\includegraphics[height=1.05em]{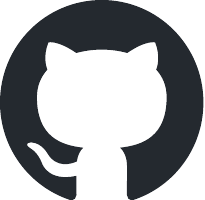}} \textbf{Github:} & \url{https://github.com/PicoTrex/Mind-Brush} \\
        \raisebox{-0.15em}{\includegraphics[height=1.05em]{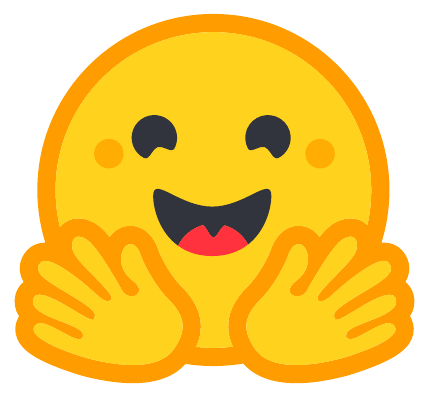}} \textbf{Dataset:} & \url{https://huggingface.co/datasets/PicoTrex/Mind-Brush} \\
        \end{tabular}
    }
{

\begin{center}
    \centering
    \captionsetup{type=figure}
    \includegraphics[width=1\textwidth]
    {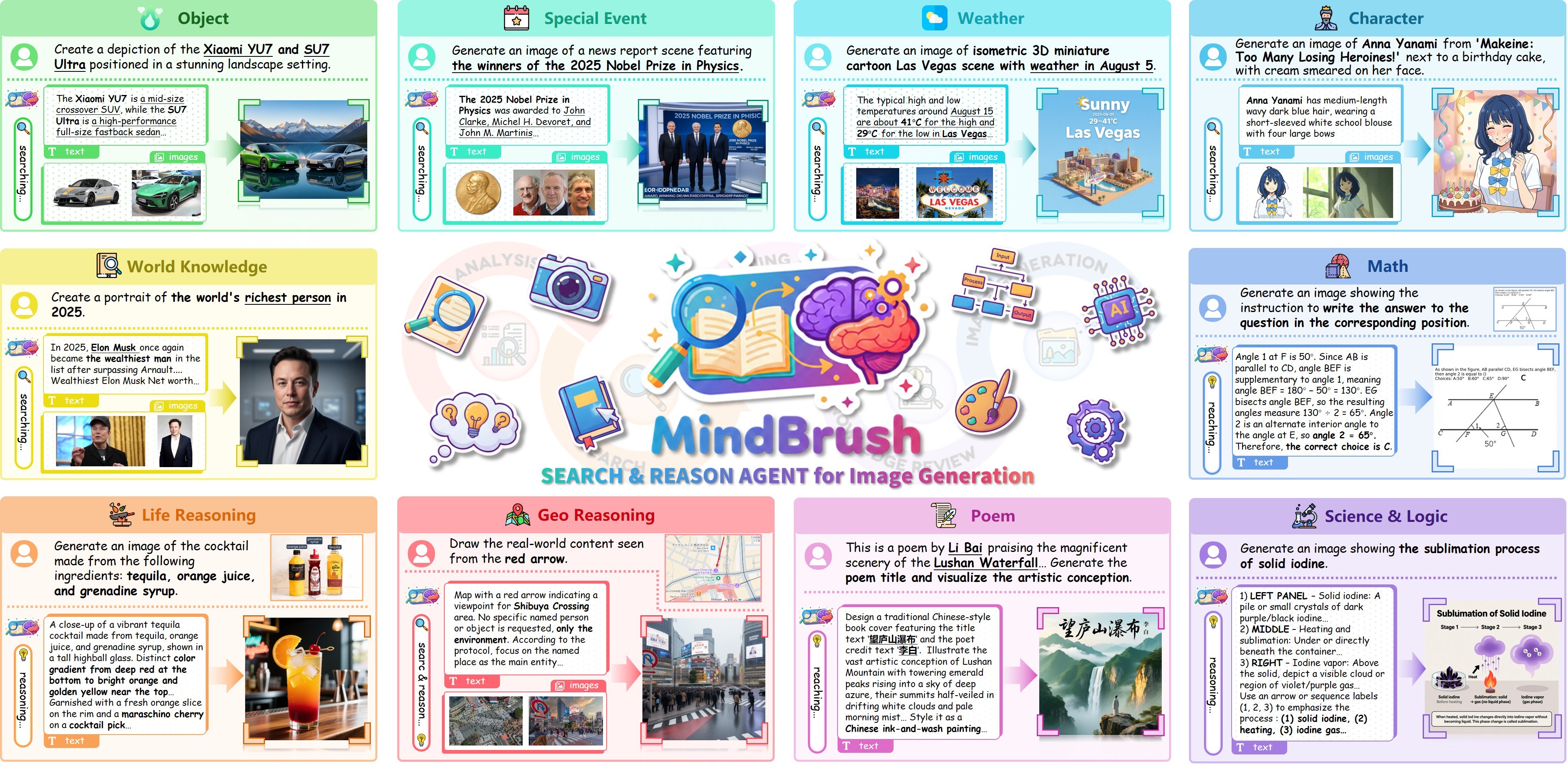}
    \captionof{figure}{We introduce \textbf{Mind-Brush}, an agentic framework that synergizes active search with explicit reasoning for image generation. By decomposing user intent, retrieving multimodal evidence, and inferring latent requirements, our agent effectively bridges the cognitive gaps and interpretative biases prevalent in existing models. Furthermore, we propose Mind-Bench, a comprehensive benchmark designed to evaluate model performance on up-to-date long-tail concepts and multimodal reasoning tasks, thereby probing the boundaries of unified understanding and generation capabilities.}
    \label{fig:teaser}
\end{center}%
\vskip 0.1in

}

]



\printAffiliationsAndNotice{\textsuperscript{*}Jun He and Junyan Ye contribute equally to this work. \\ }

\begin{abstract}

While text-to-image generation has achieved unprecedented fidelity, the vast majority of existing models function fundamentally as static text-to-pixel decoders. Consequently, they often fail to grasp implicit user intentions. Although emerging unified understanding-generation models have improved intent comprehension, they still struggle to accomplish tasks involving complex knowledge reasoning within a single model. Moreover, constrained by static internal priors, these models remain unable to adapt to the evolving dynamics of the real world. To bridge these gaps, we introduce Mind-Brush, a unified agentic framework that transforms generation into a dynamic, knowledge-driven workflow. Simulating a human-like 'think-research-create' paradigm, Mind-Brush actively retrieves multimodal evidence to ground out-of-distribution concepts and employs reasoning tools to resolve implicit visual constraints. To rigorously evaluate these capabilities, we propose Mind-Bench, a comprehensive benchmark comprising 500 distinct samples spanning real-time news, emerging concepts, and domains such as mathematical and Geo-Reasoning. Extensive experiments demonstrate that Mind-Brush significantly enhances the capabilities of unified models, realizing a zero-to-one capability leap for the Qwen-Image baseline on Mind-Bench, while achieving superior results on established benchmarks like WISE and RISE. 

\end{abstract}

\section{Introduction}

Recent advancements in image generation have democratized visual creation, enabling the seamless translation of imagination into high-fidelity imagery~\cite{rombach2022high, esser2024scaling, gong2025seedream}. However, fundamentally, the vast majority of existing models function primarily as static text-to-pixel decoders~\cite{podell2023sdxl, cai2025z}. Confined to mapping explicit user instructions to pixels, these models often fail to grasp implicit, high-level user intentions, thereby diverging significantly from the human artistic creation process. While emerging unified multimodal understanding-generation models, such as GPT-Image~\cite{openai2024gptimage1} and Bagel~\cite{deng2025bagel}, demonstrate promising capabilities in comprehending user intent and incorporating world knowledge, their performance remains constrained in tasks demanding complex mathematical or knowledge-intensive reasoning. This limitation suggests that monolithic architectures may struggle to encompass the full spectrum of capabilities required for such intricate, end-to-end tasks.

Moreover, constrained by the temporal knowledge cutoff inherent in pre-training data, the cognitive boundaries of current image generation models remain static. Consequently, they struggle to adapt to the evolving dynamics of the real world, resulting in significant capability gaps when handling real-time news or novel IP concepts~\cite{son2025world,li2025ia}. In the realm of Large Language Models (LLMs), researchers have successfully transcended these boundaries by integrating retrieval capabilities through agentic designs~\cite{chen2024mindsearch, yu2024auto}, as exemplified by Search-o1~\cite{li2025search}. While recent proprietary models such as Nano Banana Pro~\cite{team2025gemini} and FLUX-2 Max~\cite{blackforestlabs2026flux2pro} have demonstrated integrated search and reasoning capabilities, a significant gap persists within the open-source community. Specifically, there is a notable absence of open-source models capable of interacting with the open world, performing complex reasoning, and executing active planning.

Recently, agentic approaches for image generation have emerged~\cite{jiang2026genagent}, such as T2I-Copilot~\cite{chen2025t2i} and PromptSculptor~\cite{xiang2025promptsculptor}, which focus on elaborating concise instructions into detail-rich descriptions. Think-Then-Gen~\cite{kou2026think} advances this further by leveraging LLMs to decompose user queries into sequential drawing steps. Nevertheless, these efforts remain largely confined to standard T2I benchmarks like GenEval++~\cite{ye2025echo}, prioritizing prompt refinement over intricate cognitive tasks such as mathematical derivation or commonsense reasoning. Crucially, due to the absence of external tools, their reasoning relies solely on internalized knowledge. Consequently, these methods falter when tasks demand factual verification of real-time events or evolving contexts.

To bridge this gap, we introduce \textbf{Mind-Brush}, a unified agentic framework that shifts image generation from a static mapping paradigm to a dynamic, knowledge-driven workflow. Rather than treating generation as single-step inference, Mind-Brush orchestrates a cognitive process: it actively retrieves multimodal evidence to ground out-of-distribution concepts and employs logical reasoning to deduce implicit visual constraints, thereby realizing the unification of agentic understanding and generation. By effectively simulating the human artist's ``Think-Research-Create'' workflow, Mind-Brush enables high-fidelity generation requiring real-time knowledge and handles tasks involving complex reasoning.

Furthermore, existing image generation benchmarks, such as GenEval~\cite{ghosh2023geneval} and ImgEdit~\cite{ye2025imgedit}, primarily prioritize the evaluation of instruction following. While datasets like WISE~\cite{niu2025wise} and RISE~\cite{zhao2025envisioning} extend this scope to probe internal knowledge recall and rudimentary reasoning, they fall short of evaluating capabilities that require active information retrieval and complex reasoning. To address this limitation, we introduce \textbf{Mind-Bench}, a comprehensive benchmark specifically designed to assess generative performance under conditions necessitating complex reasoning and external knowledge acquisition. As illustrated in Figure~\ref{fig:teaser}, Mind-Bench encompasses 500 distinct samples across 10 diverse categories, spanning challenging scenarios from real-time news and emerging IP concepts to complex mathematical and geographical reasoning. This benchmark fills a critical void in evaluating image synthesis tasks that demand real-time knowledge and deep reasoning, particularly for unified understanding-generation models.

Our main contributions are summarized as follows:
\begin{enumerate}
    \item We propose \textbf{Mind-Brush}, a novel agentic framework that unifies intent analysis, multi-modal search, and knowledge reasoning to enable a 'think-research-create' paradigm for image generation.
    
    \item We propose \textbf{Mind-Bench}, a benchmark tailored to evaluate generative capabilities involving dynamic external knowledge and complex reasoning. Experimental results reveal critical limitations in current unified multimodal models regarding real-time awareness and logical deduction.
    
    \item Mind-Brush substantially elevates the accuracy of the Qwen-Image baseline from 0.02 to 0.31 on Mind-Bench, while significantly outperforming existing baselines on established benchmarks including the knowledge-driven WISE (+25.8 \% WiScore) and reasoning-driven RISEBench (+27.3 \% Accuracy).
    
\end{enumerate}
\section{Related Work}

\subsection{Agent for Image Generation}

Multimodal LLMs are increasingly serving as agentic decision-makers to align vague user intents with precise image synthesis\cite{yan2025unified, zhang2025unified, su2025thinking}. A primary stream focuses on prompt optimization: T2I-Copilot \cite{chen2025t2i} and PromptSculptor \cite{xiang2025promptsculptor} employ multi-agent collaboration to refine concise instructions into detailed descriptions, while ImAgent introduces test-time policy scaling for semantic alignment. Parallel works target precise control: MCCD \cite{li2025mccd} utilizes agents to decouple multi-object attributes, while AgentStory \cite{zhou2025agentstory} and CREA\cite{venkatesh2025crea} ensure narrative consistency and creative editing, respectively. To transcend static knowledge boundaries, Think-Then-Generate introduces a "think-then-generate" paradigm to explicitize internal visual logic. Concurrently, World-to-Image \cite{son2025world} and IA-T2I \cite{li2025ia} attempt to employ simple image retrieval as visual cues to supplement out-of-distribution (OOD) concepts. However, current solutions remain fragmented. Reasoning-based methods are confined by closed training data, failing on real-time events; conversely, retrieval-based approaches often treat external evidence as shallow visual cues without deep logical integration. A unified workflow synergizing active multimodal search with explicit reasoning remains absent, limiting performance on complex, knowledge-intensive tasks.

\subsection{Image Generation Model }

The ultimate goal of Unified Multi-Modal Models (UMMs) is to consolidate cross-modal understanding and generation within a single architecture. Pioneering works (e.g., Chameleon \cite{team2024chameleon}, Emu3 \cite{wang2024emu3}) integrated image generation into the LLM paradigm via visual signal discretization. However, their reliance on VQ-VAE \cite{van2017neural} introduced lossy compression, which fundamentally restricted generation fidelity. Subsequent research (e.g., Transfusion \cite{zhou2024transfusion}, Show-o \cite{xie2024show}) attempted to mitigate this by unifying autoregressive text prediction and bidirectional image diffusion within a shared Transformer backbone. Yet, this approach triggered irreconcilable modal conflicts. Even with recent architectural innovations—such as Mixture-of-Tokens (MoT) or Mixture-of-Experts (MoE) employed in Bagel and OneCat—balancing the dual objectives of understanding and generation remains a formidable challenge. Consequently, state-of-the-art methods (e.g., OmniGen2 \cite{wu2025omnigen2}, BLIP-o3 \cite{chen2025blip3}) have shifted toward a decoupled strategy, utilizing powerful multimodal large language models (MLLMs) to guide external diffusion heads, thereby achieving superior performance.

\subsection{Image Generation Benchmarks}

Evaluating Unified Multi-Modal Models (UMMs) necessitates a comprehensive assessment of both comprehension and synthesis. Mainstream benchmarks primarily assess text-image alignment and instruction following capabilities. For instance, GenEval \cite{ghosh2023geneval} evaluates compositional integrity, quantifying the model's ability to bind attributes (e.g., counts, positions) explicitly stated in prompts. However, these benchmarks are largely confined to shallow, explicit text comprehension, neglecting deeper conceptual or reasoning capabilities. To address this, subsequent benchmarks incorporate extensive world knowledge. WISE \cite{niu2025wise} and PhyBench \cite{meng2024phybench} assess domain-specific knowledge (e.g., culture, physics), while RISEBench \cite{zhao2025envisioning} focuses on logical reasoning, evaluating the translation of causal and spatio-temporal semantics into visual representations. Nevertheless, existing methods predominantly assess Internalized Parametric Memory and there is a notable absence of benchmarks evaluating multimodal reasoning or Out-of-Distribution (OOD) concepts, failing to distinguish whether a model is merely retrieving stored static knowledge or actively performing reasoning for real-time scenarios.

\begin{figure*}
    \centering
    \includegraphics[width=\linewidth]{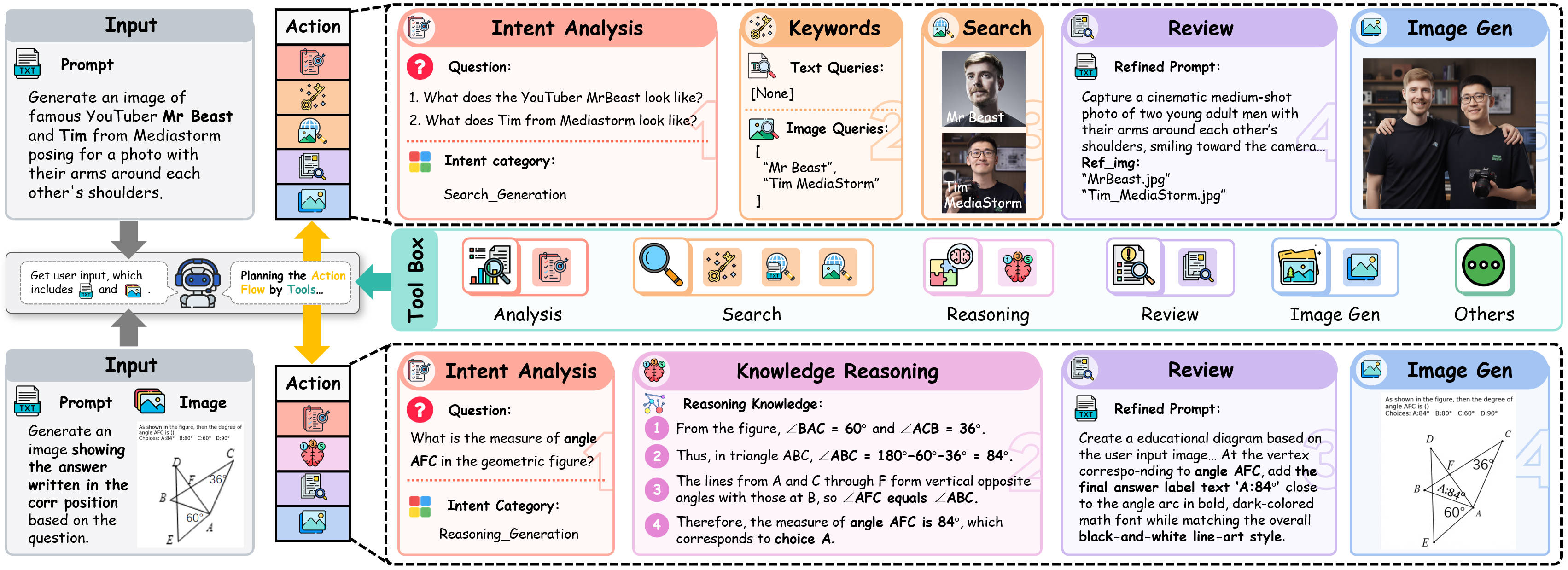}
    \caption{The overall framework of \textbf{Mind-Brush}. The user input first undergoes intent decomposition to identify potential knowledge deficits and formulate a question list. Based on specific requirements, the system dynamically executes specialized tools—such as utilizing active search or logical reasoning—to effectively bridge cognitive gaps. Finally, the consolidated evidence is organized into a final instruction via a concept review process to guide precise image generation, ensuring alignment with the user's authentic intent.} 

    \label{fig:framework_overview}
\end{figure*}

\section{Mind-Brush}
\label{sec:mind_brush}

\subsection{Problem Formulation}
\label{sec:problem_formulation}

We formalize the inference workflow of Mind-Brush as a Hierarchical Sequential Decision-Making Process, defined by the tuple $\mathcal{M} = \langle \mathcal{S}, \mathcal{A}, \pi, \mathcal{E} \rangle$. This framework generates a structured cognitive trajectory to bridge the gap between abstract intent and visual realization.

\begin{itemize}
    \item \textbf{Cognitive State ($\mathcal{S}$):} Let $s_t = \{I, I_{img}, \mathcal{E}_t\}$ denote the state at step $t$. It encapsulates the original user inputs (instruction $I$ and optional reference image $I_{img}$) and the dynamic evidence buffer $\mathcal{E}_t$, which accumulates retrieved knowledge and reasoning chains.
    
    \item \textbf{Action Space ($\mathcal{A}$):} The set of operators available to the agent. We distinguish between the Meta-Action $a_{plan}$ (Cognitive Gap Detection), which identifies cognitive gaps $Q_{gap}$, and \textit{Execution Actions} $a_{exec} \in \{a_{search}, a_{reason}\}$, which actively acquire multimodal evidence.
    
    \item \textbf{Execution Policy ($\pi$):} The Intent Analysis module functions as a high-level policy $\pi(a_{plan} | s_0)$. It assesses the initial state to formulate a deterministic execution path based on the identified $Q_{gap}$.
\end{itemize}

The inference process evolves as a context-aware trajectory. As shown in Figure~\ref{fig:framework_overview}, the system does not follow a rigid workflow; instead, it dynamically adapts to the user's request. By evaluating the specific nature of cognitive gaps in the initial state such as factual deficits or logical conflicts, the planner infers the optimal structure for evidence accumulation, routing the execution through specialized \textit{Search} or \textit{Reasoning} branches. This effectively aligns the inference computation with the intrinsic complexity of the user intent.

Ultimately, our objective is to generate the optimal target image $x^*$ based on the final converged state $s_T$. This state contains the consolidated Master Prompt $P_{master}$ and verified visual references $\mathcal{I}_{ref}$, transforming static generation into a dynamic explicit evidence accumulation process.

\subsection{Cognitive Gap Detection}

User instructions often contain implicit constraints and long-tail concepts exceeding the model's parametric knowledge boundaries. To address this, we introduce the Cognitive Gap Detection strategy, integrated within the \textbf{Intent Analysis Agent} ($\mathcal{A}_{intent}$) as a meta-planner, to bridge this cognitive divide. Specifically, it maps the text instruction $I$ and optional image $I_{img}$ into a structured semantic space via the \textbf{5W1H} (\textit{What}, \textit{When}, \textit{Where}, \textit{Why}, \textit{Who}, and \textit{How})  paradigm~\cite{cao20245w1hextractionlargelanguage}, establishing a multimodal ``Ground Truth'' to determine signal dominance. Subsequently, the module executes a rigorous gap analysis by detecting specific entities or logical dependencies that require external verification. Information absent from internal knowledge is formalized into a set of explicit atomic questions, denoted as $Q_{gap}$. Based on the composition of $Q_{gap}$, the system instantiates a dynamic execution policy $\pi$, routing the workflow to the appropriate factual grounding or logical reasoning branch defined in the action space.

\subsection{Adaptive Knowledge Completion}
\label{sec:knowledge_completion}

To bridge the identified cognitive gaps, Mind-Brush employs a Internal Logical Derivation mechanism. Unlike rigid single-pathway systems, the execution policy $\pi$ flexibly composes the retrieval and reasoning tools based on the complexity of $Q_{gap}$. 

\textbf{External Knowledge Anchoring.}
For gaps involving OOD entities or dynamic events, the framework activates the \textbf{Cognition Search Agent} ($\mathcal{A}_{search}$). It first utilizes a Keyword Generator to synthesize the user's multimodal inputs ($I, I_{img}$) and the identified gaps $Q_{gap}$, producing precise textual queries $Q_{txt}$ and initial visual queries $Q_{img}$. Upon retrieving factual documents $\mathcal{T}_{ref}$ from open-world knowledge bases, the system performs a dual-update operation:
\begin{equation}
    I' = \text{Inject}(I, \mathcal{T}_{ref}), Q'_{img} = \text{Calibrate}(Q_{img}, \mathcal{T}_{ref})
\end{equation}
where the retrieved concepts are injected back into the user instruction ($I'$) to update the textual context, while simultaneously calibrating visual queries ($Q'_{img}$) to ensure that subsequently retrieved reference images $\mathcal{I}_{ref}$ align with validated facts.

\textbf{Internal Logical Derivation.}
For gaps requiring complex deduction—such as solving mathematical problems in $I_{img}$ or inferring spatial relations from retrieved data—the system triggers the CoT \textbf{Knowledge Reasoning Agent} ($\mathcal{A}_{reasoning}$). This engine functions as a logic processor that ingests the user instruction, the input image, and crucially, the \textit{accumulated search evidence} ($\mathcal{E}_{search} = \mathcal{T}_{ref} \cup \mathcal{I}_{ref}$). It performs multi-step reasoning to resolve implicit conflicts or interpret retrieved visual data, producing explicit conclusions $\mathcal{R}_{cot}$. 

The final evidence set $\mathcal{E} = \mathcal{E}_{search} \cup \mathcal{R}_{cot}$ forms a comprehensive, logically consistent cognitive context for generation.

\subsection{Constrained Generation}
\label{sec:constrained_generation}

The accumulation of external information introduces the risk of redundancy or irrelevance. Therefore, the final phase focuses on Information Consolidation and Conditional Synthesis. First, the \textbf{Concept Review Agent} ($\mathcal{A}_{review}$) serves as a consolidation mechanism to filter noise from the disjointed evidence stream $\mathcal{E}$. It synthesizes the verified facts and logical conclusions with the user's original creative intent, rewriting them into a structured Master Prompt $P_{master}$. This prompt explicitly articulates visual attributes that were previously implicit or unknown. Subsequently, the \textbf{Unified Image Generation Agent} ($\mathcal{A}_{generation}$) executes the visual synthesis. Distinct from standard T2I models, $\mathcal{A}_{generation}$ is conditioned on both the text-aligned $P_{master}$ and adaptive visual cues $V_{in}$. Specifically, based on user intent, the mechanism dynamically selects between generation and editing modes to determine the visual conditioning source $V_{in}$ (i.e., from $\mathcal{I}_{ref}$ or $I_{img}$). These constraints effectively guide the model to achieve high fidelity to the user's creative vision while strictly adhering to the factual and logical boundaries established during the knowledge acquisition phase.

\section{Mind-Bench}
\label{sec:mind_bench}

\subsection{Motivation and Task Definition}
\label{sec:task_definition}

Complex image generation transcends simple text-to-pixel translation, necessitating a ``Research-then-Create'' paradigm akin to human artistry. However, current evaluation benchmarks often prioritize direct generation capabilities based on static knowledge. While some extend to general world knowledge, they remain relatively simplistic, suffering from a lack of \textbf{temporal sensitivity} and \textbf{depth in multimodal reasoning}. To probe the boundaries of ``cognitive generation,'' we propose \textbf{Mind-Bench}, a comprehensive benchmark comprising 500 samples designed to objectively evaluate generation capabilities dependent on dynamic external knowledge and user intent reasoning.

To systematically assess these capabilities, we categorize Mind-Bench into two primary clusters covering 10 diverse sub-domains, as shown in Figure~\ref{fig:teaser}.

\textbf{Knowledge-Driven Tasks:} This category includes five sub-domains: \textit{Special Events} (e.g., breaking news scenes), \textit{Weather} (real-time meteorological conditions), \textit{Character} (specific IP character), \textit{Object} (long-tail artifacts), and \textit{World Knowledge} (Common sense in general situations). These tasks comprehensively evaluate the model's ability to retrieve and integrate external information for precise visual grounding. The core challenge lies in mitigating hallucinations regarding Out-of-Distribution (OOD) entities.
    
\textbf{Reasoning-Driven Tasks:} This category encompasses: \textit{Life Reasoning} (common sense inference), \textit{Geo Reasoning} (spatial and map understanding), \textit{Math} (geometric and algebraic visualization), \textit{Science \& Logic} (physical states and abstract logic), and \textit{Poem} (imagery derived from literary metaphor). The core challenge lies in the model's capacity to deduce implicit constraints from ostensibly simple instructions—determining whether the model can genuinely comprehend the latent reasoning results required for accurate generation. More details of task description can be found in the appendix \ref{sec:mind_task_details}.

\begin{figure}[t]
    \centering
    \includegraphics[width=\linewidth]{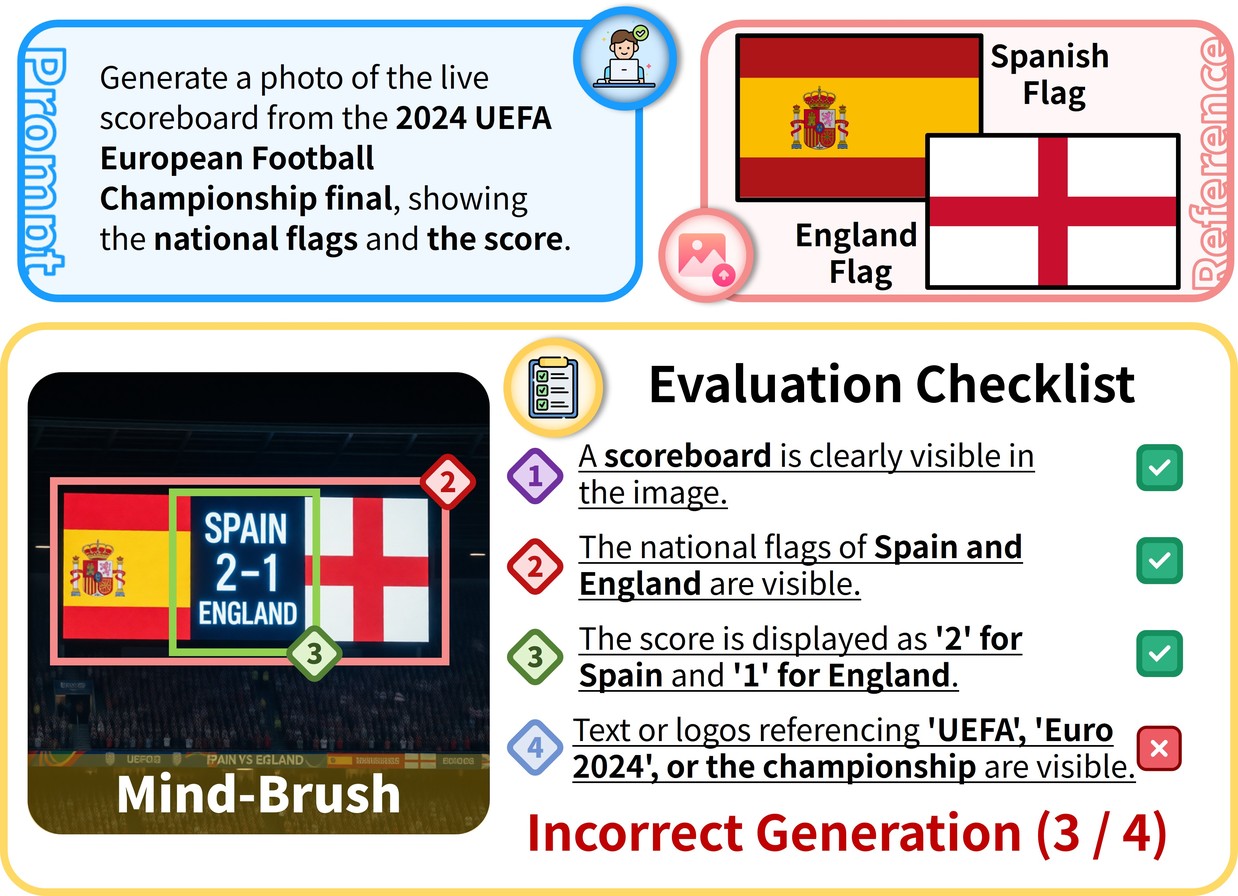}
    \caption{Overview of Checklist-based Strict Accuracy (CSA) evaluation pipeline in Mind-Bench.}
    \label{fig:mind_bench_stat}
\end{figure}

\begin{table*}[t!]
    \centering
    \caption{Quantitative comparison of different models on \textbf{Mind-Bench}. The table is divided into proprietary (top) and open-source (bottom) models. The best performing model is highlighted in \textbf{bold}. The symbol `-' indicates that the model is not applicable to I2I tasks.}
    \label{tab:comparison}
    \fontsize{8pt}{9pt}\selectfont
    \setlength{\tabcolsep}{1.9mm}
    \begin{tabular}{l|ccccc|ccccc|c}
        \toprule
        \multirow{2}{*}{\textbf{Model Name}} & \multicolumn{5}{c|}{\textbf{Knowledge-Driven}} & \multicolumn{5}{c|}{\textbf{Reasoning-Driven}} & \multirow{2}{*}{\textbf{Overall}} \\
        \cmidrule(lr){2-6} \cmidrule(lr){7-11} 
             & SE & Weather & MC & IP & WK & SL & Poem & Life Reason & GU & Math & \\
            \midrule
            GPT-Image-1\citep{openai2024gptimage1} & 0.32 & 0.06 & 0.22 & 0.02 & 0.16 & 0.32 & 0.10 & 0.24 & 0.10 & 0.12 & 0.17 \\
            GPT-Image-1.5\cite{openai2025gptimage15} & 0.36 & 0.18 & 0.22 & 0.04 & 0.30 & 0.34 & 0.08 & \textbf{0.34} & 0.10 & 0.02 & 0.21 \\
            FLUX 2 Pro\cite{blackforestlabs2026flux2pro} & 0.38 & 0.12 & 0.08 & 0.00 & 0.20 & 0.44 & 0.64 & 0.18 & 0.04 & 0.02 & 0.21 \\
            FLUX 2 Max\cite{blackforestlabs2026flux2max} & 0.44 & 0.12 & 0.10 & 0.04 & 0.38 & 0.40 & 0.50 & 0.20 & 0.02 & 0.06 & 0.23 \\
            Nano Banana\cite{deepmind2024geminiimage25} & 0.30 & 0.10 & 0.12 & 0.00 & 0.30 & 0.32 & 0.36 & 0.20 & 0.04 & 0.08 & 0.18 \\
            Nano Banana Pro\cite{deepmind2024geminiimage} & \textbf{0.50} & \textbf{0.36} & \textbf{0.40} & \textbf{0.16} & \textbf{0.56} & \textbf{0.62} & \textbf{0.68} & 0.30 & \textbf{0.16} & \textbf{0.46} & \textbf{0.41} \\
            \midrule
            SDXL\cite{podell2023sdxl} & 0.04 & 0.00 & 0.04 & 0.00 & 0.00 & 0.00 & 0.00 & - & - & - & 0.01 \\
            SD-3.5 M\cite{sd35Medium} & 0.02 & 0.00 & 0.00 & 0.00 & 0.02 & 0.00 & 0.00 & - & - & - & 0.01 \\
            SD-3.5 L\cite{sd35Large} & 0.04 & 0.00 & 0.02 & 0.00 & 0.02 & 0.00 & 0.06 & - & - & - & 0.01 \\
            FLUX 1 dev\cite{flux2024} & 0.04 & 0.00 & 0.00 & 0.00 & 0.02 & 0.02 & 0.04 & - & - & - & 0.02 \\
            FLUX 1 Kontext\cite{labs2025flux1kontextflowmatching} & 0.02 & 0.00 & 0.00 & 0.00 & 0.02 & 0.00 & 0.00 & - & - & - & 0.01 \\
            FLUX 1 Krea\cite{flux2024} & 0.04 & 0.00 & 0.04 & 0.00 & 0.02 & 0.00 & 0.02 & - & - & - & 0.02 \\
            Bagel\cite{deng2025bagel} & 0.02 & 0.00 & 0.00 & 0.00 & 0.00 & 0.02 & 0.02 & 0.02 & 0.00 & 0.08 & 0.02 \\
            Echo-4o\cite{ye2025echo} & 0.04 & 0.00 & 0.00 & 0.00 & 0.00 & 0.02 & 0.06 & 0.02 & 0.02 & 0.02 & 0.02 \\
            DraCo\cite{jiang2025draco} & 0.02 & 0.00 & 0.02 & 0.00 & 0.00 & 0.02 & 0.02 & 0.04 & 0.02 & 0.06 & 0.02 \\
            Z-Image\cite{cai2025z} & 0.02 & 0.00 & 0.08 & 0.02 & 0.00 & 0.00 & 0.00 & - & - & - & 0.02 \\
            Qwen-Image\cite{wu2025qwen} & 0.08 & 0.00 & 0.04 & 0.00 & 0.00 & 0.04 & 0.00 & 0.04 & 0.00 & 0.00 & 0.02 \\
            \rowcolor{blue!10} Mind-Brush (Ours) & \textbf{0.54} & \textbf{0.16} & \textbf{0.62} & \textbf{0.18} & \textbf{0.40} & \textbf{0.26} & \textbf{0.54} & \textbf{0.10} & \textbf{0.16} & \textbf{0.14} & \textbf{0.31} \\
            \bottomrule
        \end{tabular}
\end{table*}

\subsection{Benchmark Construction}
\label{sec:benchmark_construction}

Mind-Bench is constructed through a rigorous \textbf{Human-Machine Collaborative Pipeline} to ensure multidimensional complexity and factual reliability. 

First, distinct from random web-crawling, we recruited 6 graduate students in AI to carefully curate high-difficulty prompts. For each prompt, annotators manually collected strongly correlated \textbf{multimodal evidence} (e.g., official news reports, authoritative reference images) to establish objective factual anchors.

Subsequently, annotators utilized LLMs to generate candidate fine-grained evaluation checklist items based on the collected evidence. These items underwent strict human verification to eliminate redundancies and ensure executability. This process yields a final set of samples, each equipped with the input instruction, multimodal reference evidence, and a rigorous evaluation checklist.

\subsection{Evaluation Criterion}
\label{sec:evaluation_criterion}

Previous image generation benchmarks typically employ CLIP scores or MLLM ratings to assess generation quality across abstract perceptual dimensions. However, these metrics exhibit significant limitations in accuracy: they predominantly operate at a \textbf{coarse semantic level} (e.g., verifying the presence of a generic object) but fail to distinguish \textbf{identity-level details} or specific factual attributes.

To accurately reflect model usability in complex cognitive tasks, we propose \textbf{Checklist-based Strict Accuracy }(CSA) as the core metric, as illustrated in Figure~\ref{fig:mind_bench_stat}. This criterion employs an MLLM judge to scrutinize the generated image against the checklist under a \textbf{Holistic Pass Criterion}: a sample is deemed correct only if all sub-items are verified as ``Pass''. For a dataset with $N$ samples, the accuracy is defined as:
\begin{equation}
    Acc_{\text{CSA}} = \frac{1}{N} \sum_{i=1}^{N} \mathbb{I} \left( \prod_{j=1}^{|C_i|} \text{VQA}(I_{gen}^{(i)}, c_j^{(i)}) = 1 \right)
\end{equation}
where $\mathbb{I}(\cdot)$ is the indicator function and $\text{VQA}(I, c)$ returns 1 if image $I$ satisfies item $c$. This metric offers a rigorous evaluation, effectively penalizing generations that are partially correct but logically flawed.

\section{Experiments}

\begin{table*}[t!]
    \centering
    \caption{Quantitative comparison of different models on \textbf{WISE} and \textbf{RISEBench}. The table is divided into proprietary (top) and open-source (bottom) models. Within each group, the best performing model is highlighted in \textbf{bold}. The symbol `-' indicates that the model is not applicable to I2I tasks.}
    \label{tab:wise}
    \fontsize{8pt}{9pt}\selectfont
    \setlength{\tabcolsep}{0.9mm} 
    \begin{tabular}{l|ccccccc|cccc}
        \toprule
        \multirow{2}{*}{\textbf{Model Name}} & \multicolumn{7}{c|}{\textbf{WISE}} & \multicolumn{4}{c}{\textbf{RISEBench}} \\
        \cmidrule(lr){2-8} \cmidrule(lr){9-12}
         & Cultural & Time & Space & Bio & Phys & Chem & Overall & Instr. Reas. & App. Consis. & Vis. Plaus. & Accuracy \\
        \midrule
        GPT-Image-1\cite{openai2024gptimage1} & 0.81 & 0.71 & 0.89 & 0.83 & 0.79 & 0.74 & 0.80 & 62.8 & 80.2 & \textbf{94.9} & 28.9 \\
        Nano Banana\cite{deepmind2024geminiimage25} & \textbf{0.89} & \textbf{0.87} & \textbf{0.95} & \textbf{0.89} & \textbf{0.89} & 0.79 & \textbf{0.89} & 61.2 & \textbf{86.0} & 91.3 & 32.8 \\
        Nano Banana Pro\cite{deepmind2024geminiimage} & \textbf{0.89} & 0.80 & 0.89 & 0.88 & 0.86 & \textbf{0.85} & 0.87 & \textbf{77.0} & 85.5 & 94.4 & \textbf{47.2} \\
        \midrule
        FLUX.1-dev\cite{flux2024} & 0.48 & 0.58 & 0.62 & 0.42 & 0.51 & 0.35 & 0.50 & 26.0 & 71.6 & 85.2 & 1.9 \\
        FLUX.1-Canny\cite{flux2024} & - & - & - & - & - & - & - & 20.2 & 13.1 & 77.5 & 0.0 \\
        SD-XL-base\cite{podell2023sdxl} & 0.43 & 0.48 & 0.47 & 0.44 & 0.45 & 0.27 & 0.43 & - & - & - & - \\
        SD-3.5-medium\cite{sd35Medium} & 0.43 & 0.50 & 0.52 & 0.41 & 0.53 & 0.33 & 0.45 & - & - & - & - \\
        SD-3.5-large\cite{sd35Large} & 0.44 & 0.50 & 0.58 & 0.44 & 0.52 & 0.31 & 0.46 & - & - & - & - \\
        BAGEL (w/ CoT)\cite{deng2025bagel} & 0.76 & \textbf{0.69} & 0.75 & 0.65 & 0.75 & 0.58 & 0.70 & 45.9 & 73.8 & 80.1 & 11.9 \\
        BAGEL\cite{deng2025bagel} & 0.44 & 0.55 & 0.68 & 0.44 & 0.60 & 0.39 & 0.52 & 36.5 & 53.5 & 73.0 & 6.1 \\
        Qwen-Image\cite{wu2025qwen} & 0.62 & 0.63 & 0.77 & 0.57 & 0.75 & 0.40 & 0.62 & 49.9 & 71.0 & \textbf{91.5} & 19.4 \\
        GenAgent~\cite{jiang2026genagent} & 0.78 & 0.67 & 0.78 & \textbf{0.72} & 0.77 & 0.55 & 0.72 & - & - & - & - \\
        \rowcolor{blue!10} Mind-Brush (Ours) & \textbf{0.83} & \textbf{0.69} & \textbf{0.84} & 0.71 & \textbf{0.85} & \textbf{0.68} & \textbf{0.78} & \textbf{61.5} & \textbf{79.4} & 86.5 & \textbf{24.7} \\
        \bottomrule
    \end{tabular}%
\end{table*}

\subsection{Benchmarks and Evaluation Protocols}

To comprehensively evaluate the capabilities of current methods in understanding user intent and generating long-tail concepts, we selected three benchmarks with distinct focuses. First, our proposed \textbf{Mind-Bench} assesses generation capabilities dependent on dynamic external knowledge and multi-step reasoning using Checklist-based Strict Accuracy (CSA). We also include \textbf{WISE}, which focuses on complex semantic understanding and world knowledge integration evaluated via WiScore, and \textbf{RISEBench}, which evaluates joint text-image analysis capabilities across four reasoning dimensions (Instruction Reasoning, Appearance Consistency, Visual Plausibility, and Accuracy). Further details regarding the evaluation protocols and implementation can be found in the Appendix \ref{sec:sup_evaluation_protocols}.

\subsection{Experimental Settings}

\textbf{Baselines.} We compare Mind-Brush against current mainstream proprietary UMMs, including GPT-Image-1~\cite{openai2024gptimage1}, GPT-Image-1.5~\cite{openai2025gptimage15}, Nano Banana~\cite{deepmind2024geminiimage25}, Nano Banana Pro ~\cite{deepmind2024geminiimage}, FLUX-2 Pro~\cite{blackforestlabs2026flux2pro}, and FLUX-2 Max~\cite{blackforestlabs2026flux2max}. Additionally, we compare it with state-of-the-art (SOTA) open-source T2I models or UMMs, including FLUX.1 dev~\cite{flux2024}, FLUX 1 Kontext~\cite{labs2025flux1kontextflowmatching}, FLUX 1 Krea~\cite{flux2024}, Stable Diffusion (SD) 3.5 Large~\cite{sd35Large}, SD 3.5 Medium~\cite{sd35Medium}, SD-XL~\cite{podell2023sdxl}, Bagel~\cite{deng2025bagel}, Echo-4o~\cite{ye2025echo}, DraCo~\cite{jiang2025draco}, Z-Image~\cite{cai2025z}, GenAgent~\cite{jiang2026genagent} and Qwen-Image~\cite{wu2025qwen}. All baselines are evaluated using their official default settings.

\textbf{Implementation Details.} To ensure a fair comparison, Mind-Brush maintains consistent experimental settings across all benchmarks. We employ Qwen-Image-Edit-2512 as the image-guided T2I model and Qwen-Image as the prompt-guided T2I model. For the backbone MLLM, we uniformly utilize GPT-5.1 across all agents. Regarding search tools, we use the Google Search API for retrieval, setting the limit for text search results to 2 and truncating web content at 2000 words to ensure sufficient textual evidence without incurring excessive token costs. The limit for image search results is set to 5. All experiments involving open-source models were conducted on 8 NVIDIA A100 80G GPUs.

\begin{figure*}[t!]
    \centering
    \includegraphics[width=\linewidth]{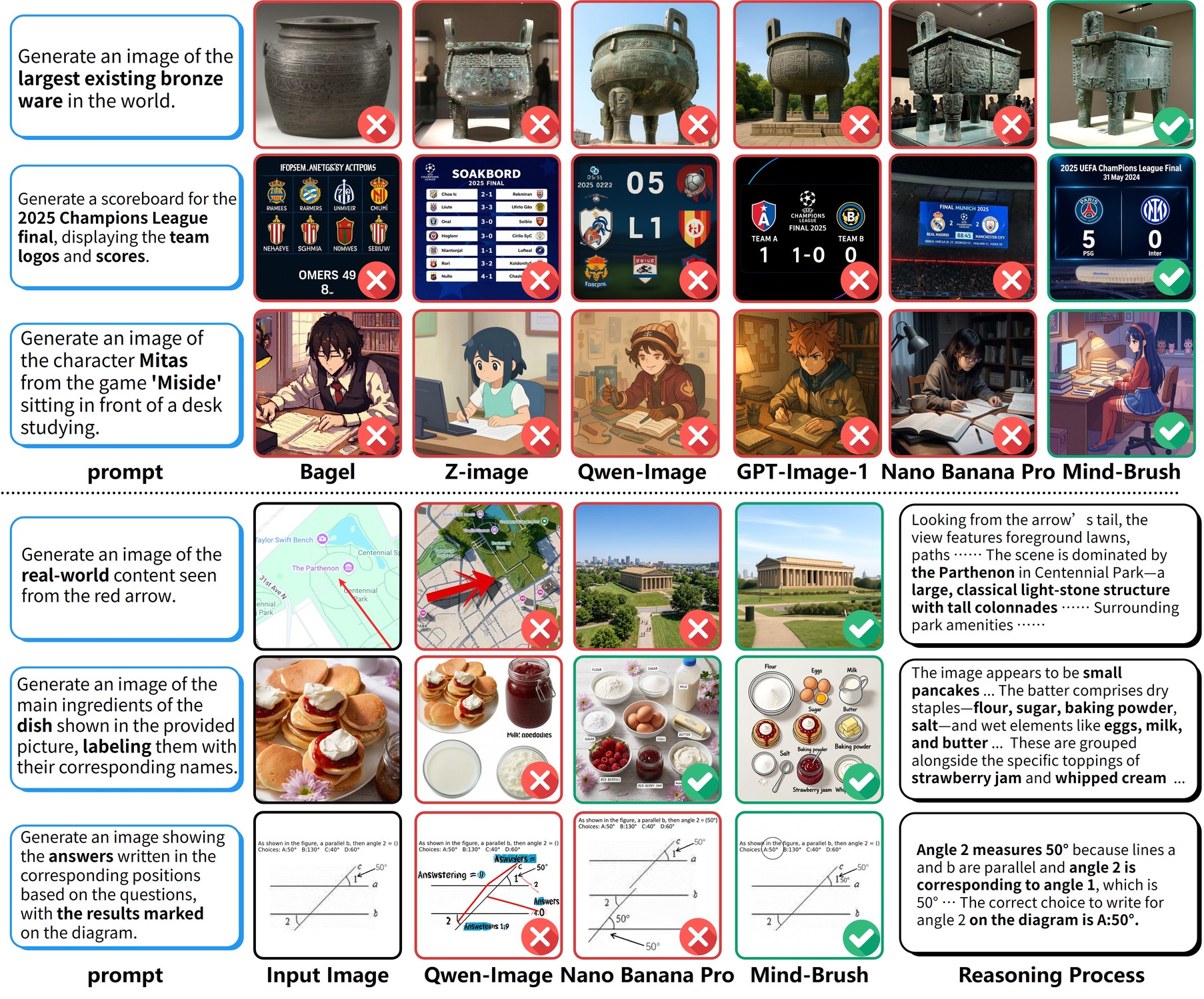}
    \caption{Qualitative Comparison of different models on Mind-Bench, including knowledge-driven (upper part) and reasoning-driven (lower part) tasks. The \textcolor{logogreen}{\textbf{green}} border indicates that the generated result matches the facts, while the \textcolor{logored}{\textbf{red}} border indicates the opposite.}
    \label{fig:qualitative}
\end{figure*}

\subsection{Comparison with State of the Art}

Table~\ref{tab:comparison} presents the quantitative comparison results on Mind-Bench. Experimental results demonstrate that Mind-Brush achieves a significant improvement in overall generation accuracy compared to open-source T2I and UMM models, surpassing SD-3.5 Large and Qwen-Image by \textbf{30.0\%} and \textbf{29.0\%}, respectively. Notably, while the baseline accuracy of Qwen-Image on Mind-Bench is considerably lower than that of proprietary methods, our Mind-Brush framework empowers these base models to match or even exceed the performance of several proprietary UMMs. Specifically, in tasks heavily reliant on search, such as \textbf{Special Events(SE)} and \textbf{Character(MC)}. In \textbf{Geo Understanding (GU)}, which demands cross-modal reasoning, Mind-Brush outperforms the vast majority of proprietary models and matches the performance of Nano Banana Pro. Furthermore, our framework achieves a \textbf{47.6\%} improvement in overall accuracy compared to GPT-Image-1.5. This validates the effectiveness of the textual and visual references provided by our retrieval and reasoning modules, achieving a unification of understanding and generation on open-source models through active knowledge acquisition and reasoning. The experimental results also indicate that Mind-Bench poses a significant challenge to current mainstream models.

\begin{table}[t]
    \centering
    \caption{Ablation Study on Knowledge and Reasoning Modules}
    \label{tab:ablation}
    \fontsize{8pt}{9pt}\selectfont
    \setlength{\tabcolsep}{1.5mm} 
    \begin{tabular}{l|ccc}
        \toprule
        Ablation Setting & \shortstack{Knowledge-\\Driven} & \shortstack{Reasoning-\\Driven} & Overall \\
        \midrule
        Baseline & 0.02 & 0.02 & 0.02 \\
        \quad + $\mathcal{A}_{reasoning}$ & 0.11 & 0.21 & 0.17 \\
        \quad + $\mathcal{A}_{search}$ & 0.30 & 0.20 & 0.25 \\
        \rowcolor{blue!10} Mind-Brush (Ours) & \textbf{0.38} & \textbf{0.24} & \textbf{0.31} \\
        \bottomrule
    \end{tabular}
\end{table}

Mind-Brush also delivers outstanding performance on \textbf{WISE}, which emphasizes world knowledge, and \textbf{RISEBench}, which focuses on reasoning logic. As shown in Table~\ref{tab:wise}, on WISE, Mind-Brush outperforms all proprietary image generation models, boosting the overall WiScore by \textbf{25.8\%} compared to Qwen-Image and reaching \textbf{0.78}, matching the performance of GPT-Image-1. On RISEBench, our method achieves a score of \textbf{61.5} in Instruction Reasoning, surpassing Nano Banana and exceeding Bagel by \textbf{68.5\%}. Furthermore, our overall accuracy approaches that of GPT-Image-1. These results further corroborate that Mind-Brush effectively integrates multimodal inputs to infer implicit user intents, thereby realizing accurate image generation.

Figure~\ref{fig:qualitative} illustrates the qualitative comparison between Mind-Brush and competing baselines on Mind-Bench. In the context of \textbf{Knowledge-Driven tasks}, Mind-Brush effectively leverages search tools to retrieve pertinent visual references, thereby achieving accurate synthesis of Out-of-Distribution concepts that baseline models fail to recognize. Regarding \textbf{Reasoning-Driven tasks}, the Knowledge Reasoning Agent dissects implicit attribute features within concise user instructions—such as spatial relationships in the output or the mathematical logic underlying input questions. This capability allows the system to effectively manifest the user's intricate intent in the generated imagery, ensuring logical coherence alongside visual fidelity.

\subsection{Ablation Study}

To validate the efficacy of the individual components within Mind-Brush, we conducted comprehensive ablation studies on the core Reasoning and Search tools using the Mind-Bench dataset. As reported in Table~\ref{tab:ablation}, the incorporation of the \textbf{Knowledge Reasoning Agent} yielded significant accuracy gains on Reasoning-Driven tasks compared to the baseline. Conversely, the \textbf{Cognition Search Agent} effectively compensated for the model's cognitive deficits regarding unknown concepts, resulting in an accuracy improvement of \textbf{0.28} on Knowledge-Driven tasks over the baseline. Furthermore, the simultaneous deployment of both agents demonstrates a strong \textbf{synergistic effect}. This combination not only bridges cognitive gaps regarding OOD concepts but also accurately deciphers the user's authentic generative intent. Consequently, the full framework achieves accuracy improvements of \textbf{0.17} and \textbf{0.06} compared to configurations using solely the Reasoning Agent or the Search Agent, respectively. More ablation study of can be found in the appendix \ref{sec:add_ablation_study}.

\section{Conclusion}

We introduced Mind-Brush, a training-free agentic framework that transforms text-to-image generation from passive decoding into an active cognitive workflow. By orchestrating intent analysis, multimodal grounding, and explicit chain-of-thought reasoning, Mind-Brush effectively bridges the gap between vague user intents and precise, factually grounded visual synthesis. To rigorously evaluate this, we established Mind-Bench, a benchmark designed to stress-test models on knowledge-intensive and reasoning-dependent tasks. Empirical results demonstrate that our framework significantly outperforms state-of-the-art models, validating the synergy of active retrieval and logical deduction. We believe this shift towards an Agentic Generative Paradigm paves the way for next-generation systems capable of complex problem-solving in visual synthesis.

\section*{Impact Statement}

This paper presents work whose goal is to advance the field of Machine Learning. There are many potential societal consequences of our work, none which we feel must be specifically highlighted here.
\bibliography{example_paper}
\bibliographystyle{icml2026}

\appendix
\newpage

\clearpage
\section*{Appendix}

\section{Additional Implementary Details of Mind-Brush}
\subsection{Workflow of Mind-Brush}

Algorithm~\ref{alg:mind_brush} formally delineates the inference workflow of the \textbf{Mind-Brush} framework. Unlike static generation paradigms, Mind-Brush operates as a dynamic, agentic cognitive system designed to align vague user intents with objective reality.

The workflow initiates with the reception of a \textbf{User Instruction} ($I_{inst}$) and an optional \textbf{User Image} ($I_{img}$), alongside the initialization of core foundation models and toolsets. In the \textbf{Intent Analysis} phase, the system structurally decomposes the input using the 5W1H paradigm to pinpoint cognitive gaps ($Q_{gap}$) and formulates a dynamic execution strategy ($\mathcal{S}_{plan}$).

Driven by this strategy, the system branches into active knowledge acquisition. If \textbf{Cognition Search} is triggered, the agent performs a two-stage retrieval process: gathering textual evidence ($\mathcal{T}_{ref}$) to ground factual concepts and subsequently refining visual queries to retrieve semantically accurate reference images ($\mathcal{I}_{ref}$). Conversely, for tasks necessitating deep logic or cross-modal understanding, the \textbf{Knowledge Reasoning} agent is activated. Crucially, this agent ingests the user image ($I_{img}$) alongside the instruction and gathered evidence, employing Chain-of-Thought (CoT) reasoning to derive explicit logical conclusions ($\mathcal{R}_{cot}$) and resolve implicit constraints.

Finally, the \textbf{Concept Review} agent acts as a convergence hub, synthesizing the user's original intent with the disjointed streams of multimodal evidence ($\mathcal{E}$) into a coherent \textbf{Master Prompt} ($P_{master}$). This prompt, enriched with factual and logical precision, guides the Image Generator ($\mathcal{G}_{\theta}$) to synthesize the final high-fidelity image ($\mathcal{I}_{final}$), ensuring rigorous alignment with both user creativity and real-world logic.

\begin{algorithm}[h!]
\caption{Inference Workflow of Mind-Brush Framework}
\label{alg:mind_brush}
\begin{algorithmic}[1]
\STATE \textbf{Input:} User Instruction $I_{inst}$, User Image $I_{img}$ (Optional)
\STATE \textbf{Initialization:} Large Language Model $\text{LLM}_{\phi}$, Image Generator $\mathcal{G}_{\theta}$, Search Tools; Evidence set $\mathcal{E} \leftarrow \emptyset$, Visual References $\mathcal{I}_{ref} \leftarrow \emptyset$

\STATE Decompose $I_{inst}$ via 5W1H paradigm and identify cognitive gaps $Q_{gap}$ ($\mathcal{A}_{intent}$)
\STATE Formulate execution strategy $\mathcal{S}_{plan}$ based on $Q_{gap}$

\IF{$\mathcal{S}_{plan}$ requires \textbf{Cognition Search}}
    \STATE Generate text queries $Q_{txt}$ and initial visual queries $Q_{v}$
    \STATE $\mathcal{T}_{ref} \leftarrow \text{Search}_{txt}(Q_{txt})_k$ 
    \STATE Update Evidence $\mathcal{E} \leftarrow \mathcal{E} \cup \mathcal{T}_{ref}$
    \STATE $Q'_{v} \leftarrow \Phi_{refine}(Q_{v}, \mathcal{T}_{ref})$ 
    \STATE $\mathcal{I}_{ref} \leftarrow \text{Search}_{img}(Q'_{v})_k$
\ENDIF

\IF{$\mathcal{S}_{plan}$ requires \textbf{Knowledge Reasoning}}
    \STATE Retrieve relevant context from $\mathcal{E}$, $I_{inst}$ and $I_{img}$
    \STATE $\mathcal{R}_{cot} \leftarrow \Phi_{reason}(I_{inst}, I_{img}, Q_{gap}, \mathcal{E})$ 
    \STATE Update Evidence $\mathcal{E} \leftarrow \mathcal{E} \cup \mathcal{R}_{cot}$
\ENDIF

\STATE Synthesize disjointed streams $\{I_{inst}, I_{img}, \mathcal{E}\}$ and resolve ambiguities ($\mathcal{A}_{review}$)
\STATE $P_{master} \leftarrow \text{Rewrite}(I_{inst}, I_{img}, \mathcal{E})$ to generate Master Prompt
\STATE $\mathcal{I}_{final} \leftarrow \mathcal{G}_{\theta}(P_{master}, \mathcal{I}_{ref})$ conditioned on visual references

\STATE \textbf{return} High-fidelity Generated Image $\mathcal{I}_{final}$
\end{algorithmic}
\end{algorithm}

\subsection{Additional Experimental Results}
Due to space limitations in the manuscript, we conducted additional evaluations of Mind-Brush on the \textbf{GenEval++}~\cite{ye2025echo} and Imagine-Bench~\cite{ye2025echo} benchmarks. 
\textbf{GenEval++} is an enhanced instruction-following benchmark that utilizes Accuracy as its metric. It increases instruction complexity and compositional difficulty across various task categories (e.g., Color, Count, Position) to mitigate metric saturation.
\textbf{Imagine-Bench} is a benchmark dedicated to creative generation, evaluating models on task categories such as Attribute Shift and Spatiotemporal Anomalies. It employs a Comprehensive Score to assess semantic understanding and visual quality in surreal fantasy scenarios.

Experimental results demonstrate that Mind-Brush achieves impressive performance on these challenging benchmarks. 
As shown in Table~\ref{tab:geneval_performance}, compared to Agentic methods such as PromptEnhancer and GenAgent, our method outperforms the best performance baseline (GenAgent) on GenEval++. 
Specifically, Mind-Brush improves Accuracy by 41.7\% on the Pos/Count task and 13.3\% on the Multi-Count task, reaching performance levels close to GPT-4o. 
Table~\ref{tab:imagine_performance} presents the performance comparison on Imagine-Bench; our results show improvements in Score of 8.2\% and 2.6\% on the Spatiotemporal and Hybridization tasks, respectively, compared to GenAgent. 
Extensive experimental results validate the superior capability of our framework in handling complex long-tail instructions and open-world creative generation tasks.

\begin{table*}[t!]
\centering
\caption{Quantitative Comparison of different methods on \textbf{GenEval++}. The best performing model (excluding close-sourece model GPT4o) is highlighted in \textbf{bold} and \underline{underline} indicates the second best.}
\label{tab:geneval_performance}
\fontsize{9pt}{12pt}\selectfont
\setlength{\tabcolsep}{1.8mm} 
\begin{tabular}{l|cccccccc}
\toprule
Method & Color & Count & Color/Count & Color/Pos & Pos/Count & Pos/Size & Multi-Count & Overall \\
\midrule
FLUX.1-dev~\cite{flux2024} & 0.400 & 0.600 & 0.250 & 0.250 & 0.075 & 0.400 & 0.300 & 0.325 \\
Qwen-Image~\cite{wu2025qwen} & \textbf{0.875} & \underline{0.725} & \underline{0.725} & 0.600 & 0.475 & \underline{0.725} & 0.550 & 0.668 \\
\midrule
\color{gray}GPT4o~\cite{achiam2023gpt} & \color{gray}0.900 & \color{gray}0.675 & \color{gray}0.725 & \color{gray}0.625 & \color{gray}0.600 & \color{gray}0.800 & \color{gray}0.850 & \color{gray}0.739 \\
Janus Pro 7B~\cite{chen2025janus} & 0.450 & 0.300 & 0.125 & 0.300 & 0.075 & 0.350 & 0.125 & 0.246 \\
T2I-R1~\cite{jiang2025t2i} & 0.675 & 0.325 & 0.200 & 0.350 & 0.075 & 0.250 & 0.300 & 0.311 \\
Bagel~\cite{deng2025bagel} & 0.325 & 0.600 & 0.250 & 0.325 & 0.250 & 0.475 & 0.375 & 0.371 \\
\midrule
PromptEnhancer~\cite{xiang2025promptsculptor} & 0.500 & 0.625 & 0.225 & 0.375 & 0.125 & 0.450 & 0.375 & 0.382 \\
ReflectionFLow~\cite{zhuo2025reflection} & 0.400 & 0.625 & 0.275 & 0.275 & 0.200 & 0.425 & 0.325 & 0.361 \\
GenAgent~\cite{jiang2026genagent} & \underline{0.775} & \textbf{0.775} & 0.650 & \textbf{0.800} & \underline{0.600} & \underline{0.725} & \underline{0.750} & \underline{0.725} \\
\midrule
\rowcolor{blue!10}Mind-Brush (Ours) & \underline{0.775} & 0.700 & \textbf{0.775} & \underline{0.750} & \textbf{0.850} & \textbf{0.775} & \textbf{0.850} & \textbf{0.782} \\
\bottomrule
\end{tabular}
\end{table*}

\begin{table*}[h]
\centering
\caption{Quantitative Comparison of different methods on \textbf{Imagine}. The best performing model (excluding close-sourece model GPT4o) is highlighted in \textbf{bold} and \underline{underline} indicates the second best.}
\label{tab:imagine_performance}
\fontsize{9pt}{12pt}\selectfont
\setlength{\tabcolsep}{2mm} 
\begin{tabular}{l|ccccc}
\toprule
Method & Attribute Shift & Spatiotemporal & Hybridization & Multi-Object & Overall \\
\midrule
FLUX.1-dev~\cite{flux2024} & 5.298 & 6.350 & 7.053 & 5.973 & 6.072 \\
Qwen-Image~\cite{wu2025qwen} & 6.771 & 7.193 & 8.130 & 7.500 & 7.329 \\
\midrule
\color{gray}GPT4o~\cite{achiam2023gpt} & \color{gray}8.540 & \color{gray}9.180 & \color{gray}8.570 & \color{gray}7.980 & \color{gray}8.560 \\
Janus Pro 7B~\cite{chen2025janus} & 5.300 & 7.280 & 6.730 & 6.040 & 6.220 \\
T2I-R1~\cite{jiang2025t2i} & 5.850 & \underline{7.700} & 7.360 & 6.680 & 6.780 \\
Bagel~\cite{deng2025bagel} & 5.370 & 6.930 & 6.500 & 6.410 & 6.200 \\
\midrule
PromptEnhancer~\cite{xiang2025promptsculptor} & 5.489 & 6.213 & 7.327 & 6.493 & 6.281 \\
GenAgent~\cite{jiang2026genagent} & \textbf{7.613} & 7.547 & \underline{8.343} & \textbf{7.763} & \underline{7.794} \\
\midrule
\rowcolor{blue!10}Mind-Brush (Ours) & \underline{7.416} & \textbf{8.167} & \textbf{8.557} & \underline{7.533} & \textbf{7.862} \\
\bottomrule
\end{tabular}
\end{table*}

\subsection{Additional Ablation Study}
\label{sec:add_ablation_study}

To rigorously evaluate the generalizability and robustness of the Mind-Brush framework, we conducted ablation studies involving diverse configurations of the Multimodal Large Language Model (MLLM) backbone and the image generation engine. The quantitative results, presented in Table~\ref{tab:model_ablation}, confirm the framework's efficacy across distinct computational regimes and disentangle the contributions of the agentic intelligence substrate versus the visual synthesis capabilities.

\textbf{Quantitative Comparison with Different MLLM Baselines.} 
To isolate the contribution of the multimodal foundation model that drives the entire agentic workflow, we evaluated the framework's performance by replacing the MLLM backbone across all constituent agents while holding the generation model constant. We first tested a fully open-source configuration utilizing \textit{Qwen3-VL-235B} as the universal backbone for all agents and \textit{Qwen-Image} as the visual generator. Despite relying on accessible open-weights models, this configuration achieves an overall CSA score of \textbf{0.24}, surpassing the proprietary baseline \textit{GPT-Image-1.5} (0.21) by a relative margin of \textbf{14.3\%}. This empirical evidence suggests that the collaborative agentic workflow can effectively compensate for the capacity gaps between open-source models and state-of-the-art proprietary baselines. Furthermore, by upgrading the framework's backbone from \textit{Qwen3-VL} to \textit{GPT-5.1} (while keeping \textit{Qwen-Image} fixed), the overall performance significantly improves from 0.24 to \textbf{0.31}, yielding a \textbf{29.2\% relative gain}. This finding is pivotal, demonstrating that the intelligence level of the underlying MLLM is the dominant factor. A more capable backbone enhances the precision of every agentic step—from evidence retrieval to constraint verification—directly translating into higher generation fidelity even without altering the image decoder.

\textbf{Quantitative Comparison with Different Image Generation Models.}
We further investigated the framework's capacity to empower and scale with different visual executors. Integrating the legacy \textit{GPT-Image-1} into the Mind-Brush framework (driven by the GPT-5.1 backbone) results in a substantial performance leap, where the overall CSA score increases from the baseline of 0.17 to \textbf{0.34}, representing a \textbf{100\% relative improvement}. This confirms that Mind-Brush serves as a powerful meta-architecture capable of unlocking the latent potential of weaker generators. Notably, under the same agentic backbone (\textit{GPT-5.1}), the \textit{GPT-Image-1} based configuration outperforms the \textit{Qwen-Image} variant. This indicates that while the agentic backbone is crucial for intent alignment and factuality, the framework effectively utilizes the superior visual synthesis capabilities of stronger generation engines, ensuring that performance scales consistently with the quality of both the intelligence substrate and the visual decoder.

\begin{table*}[t!]
    \centering
    \caption{Ablation study of different MLLM backbones and image generation model. The best performing model is highlighted in \textbf{bold}.}
    \fontsize{9pt}{12pt}\selectfont
    \setlength{\tabcolsep}{1.5mm}
    \label{tab:model_ablation}
    \begin{tabular}{c|c|ccccc|ccccc|c}
        \toprule
        \multirow{2}{*}{\textbf{MLLM model}} & \multirow{2}{*}{\textbf{Generation model}} & \multicolumn{5}{c|}{\textbf{Knowledge-Intensive}} & \multicolumn{5}{c|}{\textbf{Reasoning-Intensive}} & \multirow{2}{*}{\textbf{Overall}} \\
        \cmidrule(lr){3-7} \cmidrule(lr){8-12} 
         & & SE & Weather & MC & IP & WK & Life Reason & GU & Math & SL & Poem & \\
        \midrule
           - & GPT-Image-1 & 0.32 & 0.06 & 0.22 & 0.02 & 0.16 & 0.24 & 0.10 & 0.12 & 0.32 & 0.10 & 0.17 \\
           - & GPT-Image-1.5 & 0.36 & \underline{0.18} & 0.22 & 0.04 & 0.30 & \textbf{0.34} & 0.10 & 0.02 & 0.34 & 0.08 & 0.21 \\
           - & Nano Banana & 0.30 & 0.10 & 0.12 & 0.00 & 0.30 & 0.20 & 0.04 & 0.08 & 0.32 & 0.36 & 0.18 \\
           - & Nano Banana Pro & 0.50 & \textbf{0.36} & 0.40 & \underline{0.16} & \textbf{0.56} & \underline{0.30} & \textbf{0.16} & \textbf{0.46} & \textbf{0.62} & \textbf{0.68} & \textbf{0.41} \\
            \midrule
            \multicolumn{13}{l}{\textit{\textbf{Mind-Brush}}} \\
        \rowcolor{blue!10} Qwen3-VL-235B & Qwen-Image & 0.42 & 0.06 & 0.44 & 0.10 & 0.36 & 0.06 & \underline{0.14} & \underline{0.18} & 0.20 & 0.44 & 0.24 \\
        \rowcolor{gray!10} GPT-5.1 & Qwen-Image & \underline{0.54} & 0.16 & \textbf{0.62} & \textbf{0.18} & 0.40 & 0.10 & \textbf{0.16} & 0.14 & 0.26 & \underline{0.54} & 0.31 \\
        \rowcolor{blue!10}GPT-5.1 & GPT-Image-1   & \textbf{0.64} & \underline{0.18} & \underline{0.56} & 0.10 & \underline{0.50} & 0.28 & 0.10 & 0.06 & \underline{0.50} & 0.48 & \underline{0.34} \\
        \bottomrule
    \end{tabular}
\end{table*}

\subsection{Evaluation Protocols}
\label{sec:sup_evaluation_protocols}

To ensure a rigorous and standardized assessment of model performance across diverse cognitive dimensions, we strictly adhere to the official evaluation protocols defined in the respective benchmarks. All automated evaluations for existing benchmarks utilize the official LLM settings as the core judge to maintain consistency with prior literature.

\paragraph{WISE Evaluation.}
For the \textbf{WISE} benchmark, which necessitates deep semantic comprehension and world knowledge integration, we report the \textbf{WiScore} following the official definition. This composite metric provides a quantitative measure of knowledge-image alignment by aggregating three sub-dimensions on a discrete 3-point scale ($s \in \{0, 1, 2\}$): \textit{Consistency} ($S_{con}$), \textit{Realism} ($S_{real}$), and \textit{Aesthetic Quality} ($S_{aes}$).
Formally, the WiScore is calculated as a weighted linear combination:
\begin{equation}
    \text{WiScore} = \alpha_1 \cdot \bar{S}_{con} + \alpha_2 \cdot \bar{S}_{real} + \alpha_3 \cdot \bar{S}_{aes}
\end{equation}
The weights are rigorously calibrated to $\alpha_1=0.7$, $\alpha_2=0.2$, and $\alpha_3=0.1$. This weighting strategy prioritizes the model's fidelity to implicit semantic constraints (Consistency) while simultaneously penalizing violations of physical laws (Realism) and low artistic quality (Aesthetic), thereby offering a holistic view of generation quality grounded in world knowledge.

\paragraph{RISEBench Evaluation.}
\textbf{RISEBench} targets reasoning-informed visual editing, where precision is paramount. Following its official guidelines, we conduct a fine-grained evaluation using GPT-4.1 to score results on a scale of 1 to 5 across three critical dimensions: \textit{Instruction Reasoning} ($S_{ir}$), which measures the accurate execution of complex logical directives; \textit{Appearance Consistency} ($S_{ac}$), which assesses the preservation of task-irrelevant visual attributes; and \textit{Visual Plausibility} ($S_{vp}$), which evaluates the coherence of the edited output.
Distinct from standard aggregation methods, RISEBench imposes a strict \textit{All-or-Nothing} success criterion. A sample is deemed successfully solved if and only if it achieves perfect scores across all applicable dimensions. The final accuracy is defined as:
\begin{equation}
    \text{Acc}_{\text{RISE}} = \frac{1}{N} \sum_{i=1}^{N} \mathbb{I}\left( S_{ir}^{(i)}=5 \land S_{ac}^{(i)}=5 \land S_{vp}^{(i)}=5 \right)
\end{equation}
where $\mathbb{I}(\cdot)$ denotes the indicator function. This rigorous metric effectively penalizes partial failures, ensuring that high performance reflects robust reasoning capabilities rather than successful approximations.

\paragraph{Mind-Bench Evaluation.}
For our proposed \textbf{Mind-Bench}, to rigorously quantify the model's ability to bridge cognitive gaps without hallucination, we employ the \textbf{Checklist-based Strict Accuracy (CSA)}. To mitigate potential self-evaluation biases inherent in GPT-series models, we utilize \textbf{Gemini-3.0-Pro} as the expert evaluator.
Each test sample is associated with a human-verified set of atomic factual claims $\mathcal{C} = \{c_1, \dots, c_k\}$. The evaluator performs a binary verification ($\{0, 1\}$) for each claim. We adopt a \textit{Holistic Pass Criterion}, where a generation is considered correct only if it satisfies all constraints in the checklist:
\begin{equation}
    \text{Acc}_{\text{CSA}} = \frac{1}{N} \sum_{i=1}^{N} \mathbb{I} \left( \prod_{j=1}^{|\mathcal{C}_i|} \text{VQA}_{\mathcal{M}}(I_{gen}^{(i)}, c_j^{(i)}) = 1 \right)
\end{equation}
This stringent protocol ensures that the reported accuracy reflects comprehensive understanding and precise grounding of external knowledge, rather than superficial semantic overlap.

\begin{table*}[t]
\centering
\small
\renewcommand{\arraystretch}{1.5}
\caption{\textbf{Overview of Mind-Bench.} Task distribution, types, and per-task definitions across Knowledge and Reasoning domains.}
\label{tab:task_breakdown}

\begin{tabular}{l c l p{10.5cm}} 
\toprule
\textbf{Task} & \textbf{Number} & \textbf{Type} & \textbf{Definition} \\ \midrule

\rowcolor{gray!15} \multicolumn{4}{l}{\textbf{Knowledge-driven}} \\ 
News & 50 & T2I & Retrieve and generate images of specific news events based on the provided temporal and spatial contexts. \\
Weather & 50 & T2I & Retrieve and generate images of meteorological conditions for specific times and locations. \\
Character & 50 & T2I & Retrieve and generate images of specific personas, celebrities, or fictional characters from user input. \\
IP & 50 & T2I & Retrieve and generate images of products and artifacts associated with well-known intellectual properties. \\
World Knowledge & 50 & T2I & Retrieve and generate images corresponding to specific factual and historical information about the world. \\ \midrule

\rowcolor{gray!15} \multicolumn{4}{l}{\textbf{Reasoning-driven}} \\ 
Life Reasoning & 50 & I2I & Reason and generate images related to daily life tasks and their outcomes based on provided lifestyle imagery. \\
Geo Understanding & 50 & I2I & Reason, retrieve, and generate images of specific locations based on input map imagery and spatial contexts. \\
Math & 50 & I2I & Reason visual mathematical problems and generate images rendering the step-by-step results. \\
Science & 50 & T2I & Reason and generate images depicting scientific phenomena, physical states, or experimental processes. \\ 
Poem & 50 & T2I & Reason and generate visual scenes that embody the specific imagery and metaphors of poems given the poet and emotion. \\
\bottomrule
\end{tabular}
\end{table*}

\begin{table*}[h]
    \centering
    \caption{Comparison between Mind-Bench and existing T2I benchmarks. }
    \label{tab:benchmark_comparison}
    
    \fontsize{9pt}{12pt}\selectfont
    \begin{tabular}{lccccc}
        \toprule
        \textbf{Benchmark} & \textbf{Up-to-date} & \textbf{Reasoning modality} & \textbf{Sample Num} & \textbf{Task Num} & \textbf{Metric Type} \\
        \midrule
        GenEval~\cite{ghosh2023geneval} & No & Text & $\sim$550 & 6 & Scoring \\
        GenEval++~\cite{ye2025echo}& No & Text & $\sim$280 & 7 & Accuracy \\
        WISE~\cite{niu2025wise} & No & Text & 1,000 & 25 & Scoring \\
        T2I-ReasonBench~\cite{sun2025t2i} & No & Text + Image & 800 & 4 & Scoring \\
        RISEBench~\cite{zhao2025envisioning} & No & Text + Image & 360 & 4 & Scoring + Accuracy \\
        \midrule
        \textbf{Mind-Bench (Ours)} & \textbf{Yes } & \textbf{Text + Image} & \textbf{500} & \textbf{10} & \textbf{Accuracy (CSA)} \\
        \bottomrule
    \end{tabular}%
\end{table*}

\section{Additional Details of Mind-Bench}
\label{sec:mind_bench_details}

\subsection{Additional task description of Mind-Bench}
\label{sec:mind_task_details}

Due to space limitations in the main manuscript, we provide a comprehensive specification of the task taxonomy within Mind-Bench in this section. Table~\ref{tab:task_breakdown} presents a detailed overview, outlining the sample distribution, input modalities, and precise definitions for each of the 10 distinct task categories, structured across Knowledge-driven and Reasoning-driven domains to fully elucidate the benchmark's evaluation scope.

\subsection{Comparison with Existing Benchmarks}

To contextualize the contributions of Mind-Bench, Table~\ref{tab:benchmark_comparison} presents a comprehensive comparison with existing T2I benchmarks, highlighting our distinct advantages in temporality, modality, and evaluation rigor. Uniquely, Mind-Bench serves as the sole platform for assessing \textbf{real-time information retrieval}, diverging from traditional benchmarks like GenEval and WISE that rely on static, frozen knowledge distributions to explicitly target dynamic concepts. Beyond simple text alignment, it incorporates \textbf{multimodal contexts} to evaluate complex reasoning, matching the depth of specialized benchmarks like T2I-ReasonBench while offering superior breadth across \textbf{10 distinct task categories}. Furthermore, by adopting the \textbf{Checklist-based Strict Accuracy (CSA)} metric instead of subjective scalar scoring, Mind-Bench ensures a precise, objective standard for validating the capabilities of agentic generation systems.

\subsection{Data Sources and Copyright Statement}

The construction of Mind-Bench strictly adheres to the copyright policies of its data sources. Data is primarily derived from publicly available academic datasets or public information sources permitting non-commercial use:

\begin{enumerate}

    \item \textbf{Wikipedia:} Reference images and textual descriptions for tasks involving News, Historical Events, Specific Characters, Trendy IPs, World Knowledge, and Geographical Understanding are primarily sourced from Wikipedia\footnote{\url{https://www.wikipedia.org/}}. These contents are utilized under the \textbf{Creative Commons Attribution-ShareAlike 3.0 Unported License (CC BY-SA 3.0)}.

    \item \textbf{Historical Weather Data:} Data for Weather tasks, including temperature and historical weather conditions, is sourced from \texttt{world-weather.info}\footnote{\url{https://world-weather.info}}.

    \item \textbf{Life Reasoning Data:} Image data for life reasoning tasks is curated from \textit{recipetineats}\footnote{\url{https://www.recipetineats.com}}.

    \item \textbf{Mathematical Data:} Images and queries for Math tasks are primarily adapted from the \textbf{MathVerse} image generation benchmark~\cite{zhang2024mathverse}, adhering to its academic citation license.
\end{enumerate}

\textbf{Redistribution Policy:} We prioritize respecting the intellectual property rights of all data owners. For content licensed under compatible open-source agreements (e.g., CC BY-SA), we include the data directly in our release. For sources that restrict secondary redistribution, our benchmark release will provide metadata and direct URLs to the original content, allowing users to download the data independently in compliance with the respective terms of service.

\section{Additional Visualization of Inference Process}
\label{sec:more_vis}

\begin{figure*}[t!]
    \centering
    \includegraphics[width=0.8\linewidth]{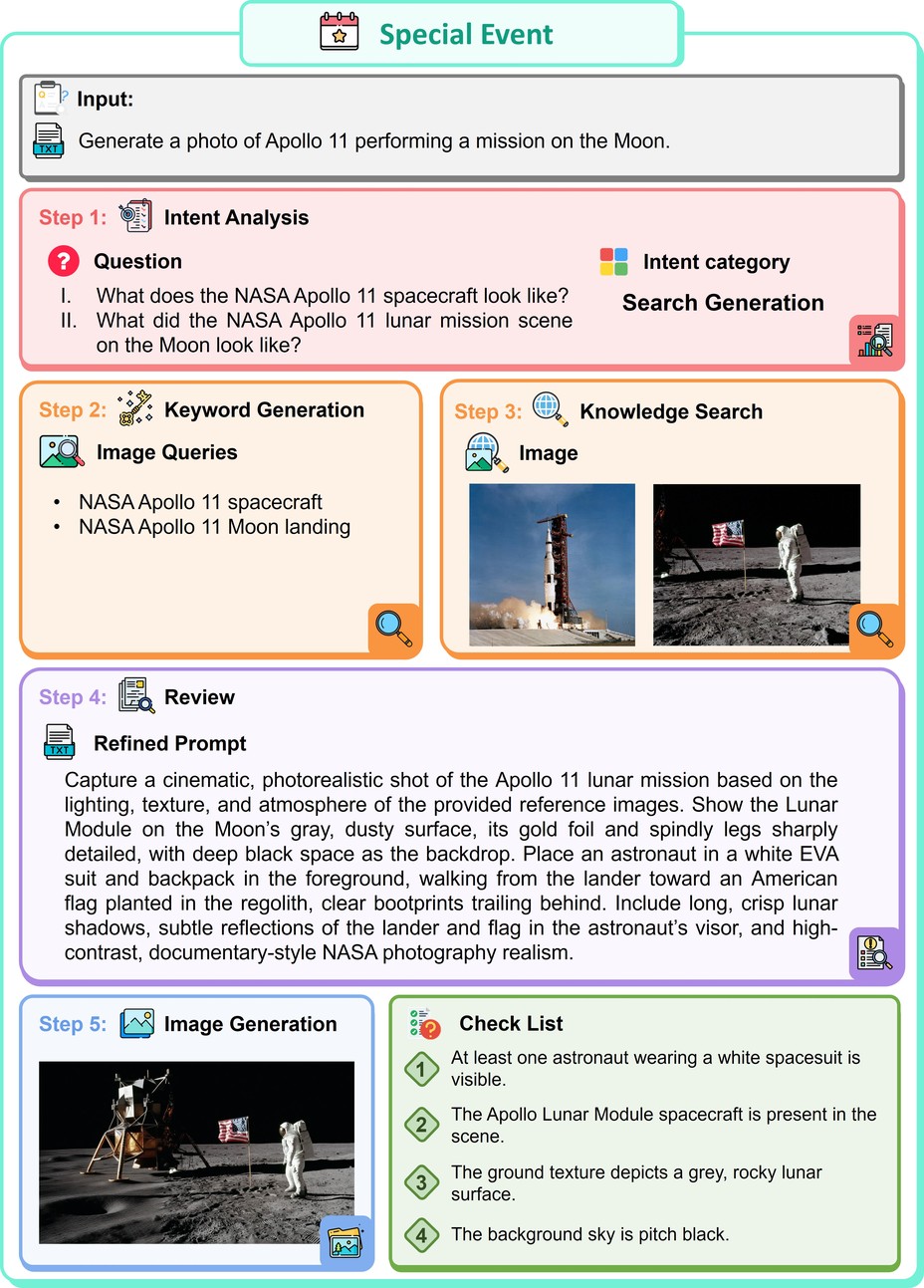}
    \caption{A generation process of Mind-Brush in Special Events task of Mind-Bench.}
    \label{fig:sup_event1}
\end{figure*}

\begin{figure*}[t!]
    \centering
    \includegraphics[width=0.75\linewidth]{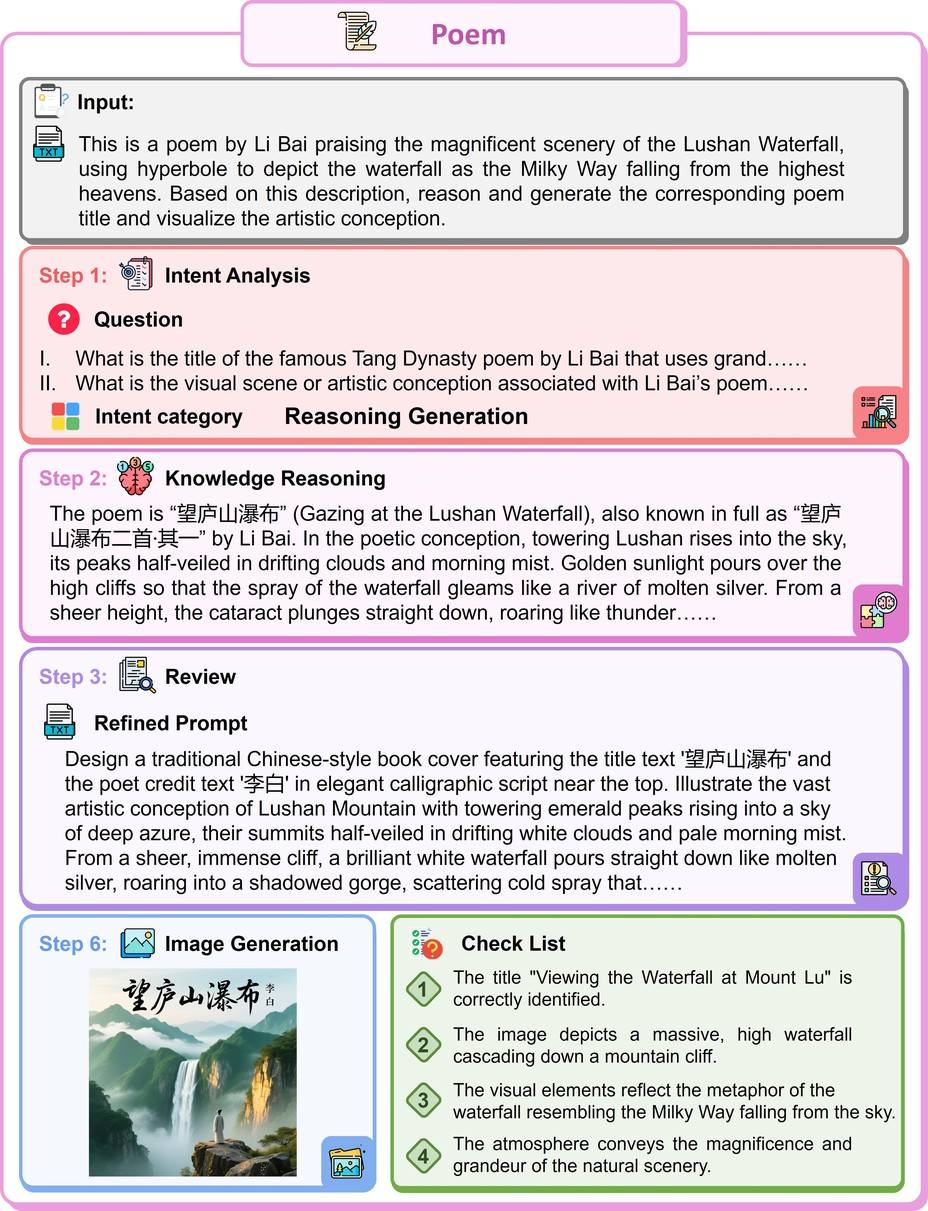}
    \caption{A generation process of Mind-Brush in Poem task of Mind-Bench.}
    \label{fig:sup_Poem2}
\end{figure*}

To provide a more intuitive understanding of our framework, we present additional qualitative examples (from fig. \ref{fig:sup_event1} to fig. \ref{fig:sup_Poem2}) that visualize the complete step-by-step cognitive trajectory of Mind-Brush, ranging from initial intent analysis to the final image synthesis across diverse scenarios.

\newpage

\begin{figure*}[t!]
    \centering
    \includegraphics[width=0.75\linewidth]{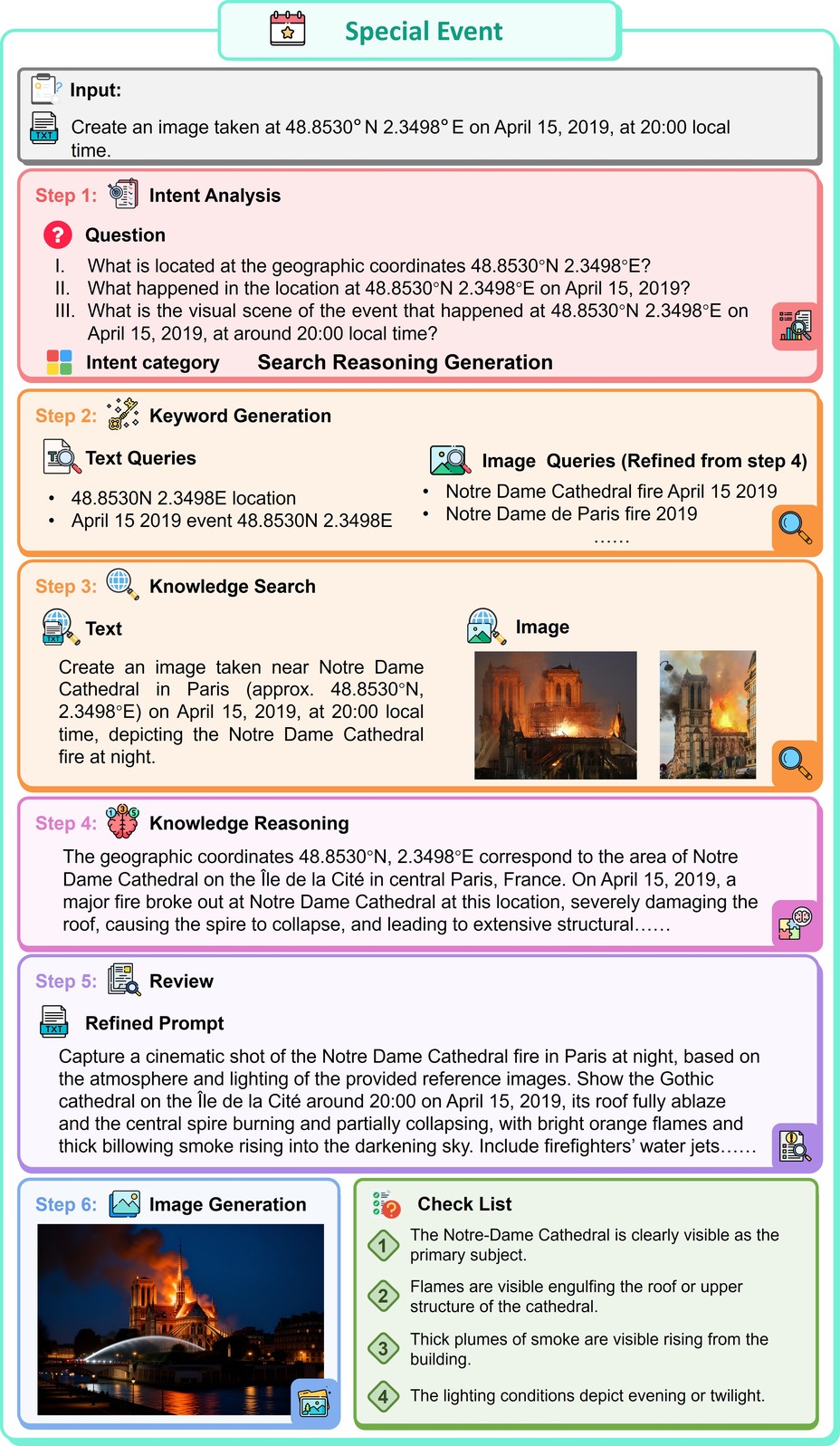}
    \caption{A generation process of Mind-Brush in Special Events task of Mind-Bench.}
    \label{fig:sup_event2}
\end{figure*}

\begin{figure*}[t!]
    \centering
    \includegraphics[width=0.75\linewidth]{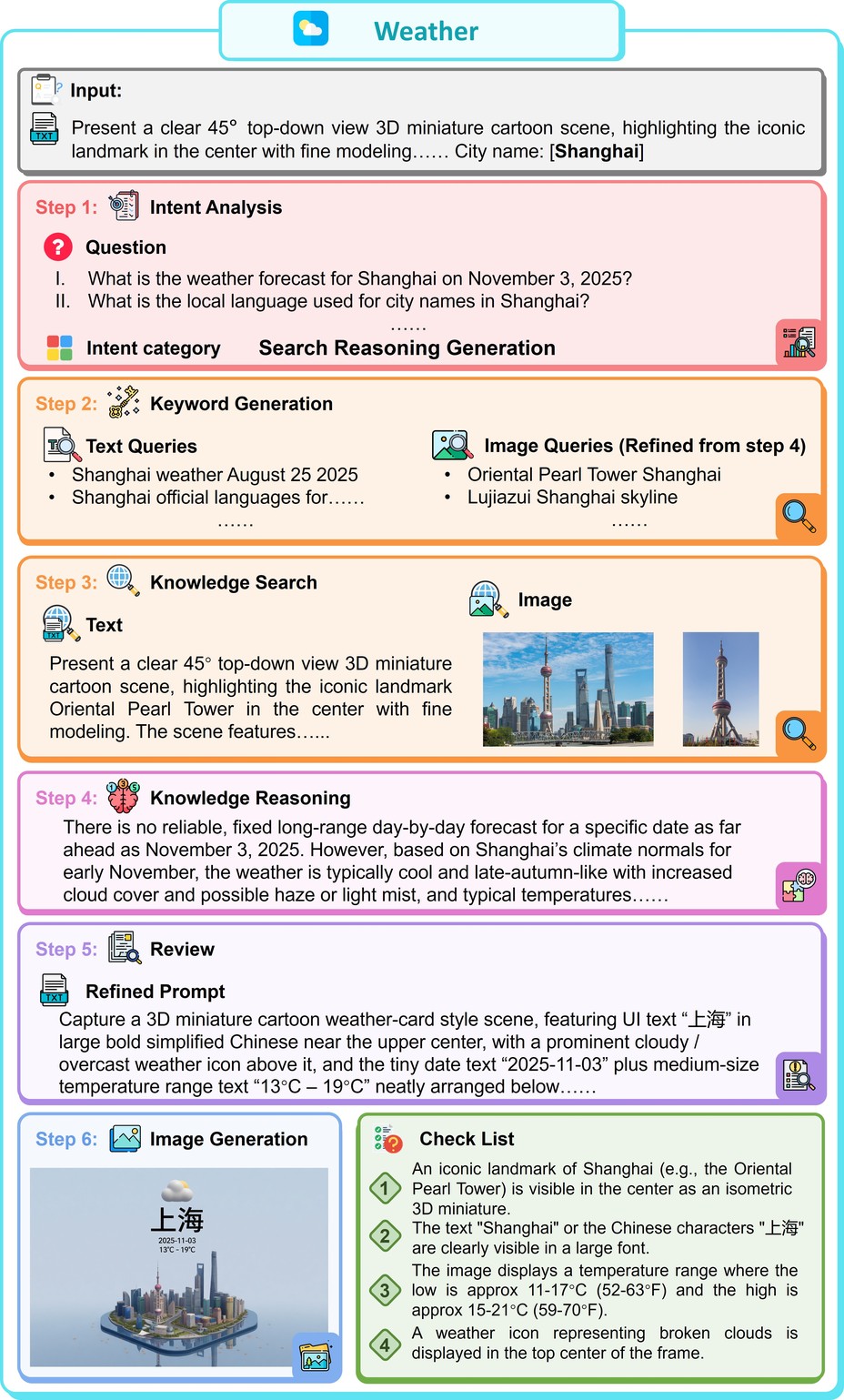}
    \caption{A generation process of Mind-Brush in Weather task of Mind-Bench.}
    \label{fig:sup_weather1}
\end{figure*}

\begin{figure*}[t!]
    \centering
    \includegraphics[width=0.75\linewidth]{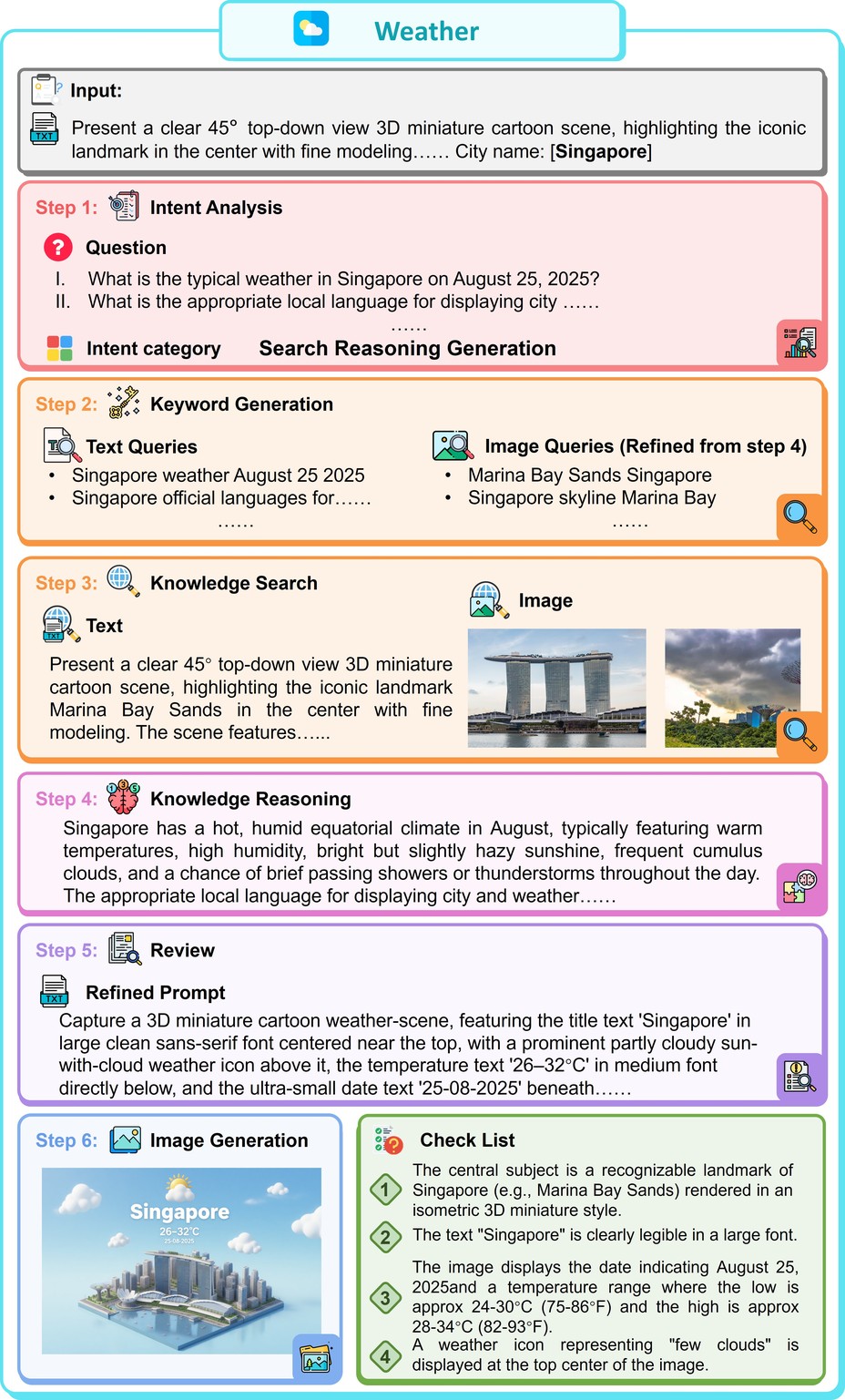}
    \caption{A generation process of Mind-Brush in Weather task of Mind-Bench.}
    \label{fig:sup_weather2}
\end{figure*}

\begin{figure*}[t!]
    \centering
    \includegraphics[width=0.75\linewidth]{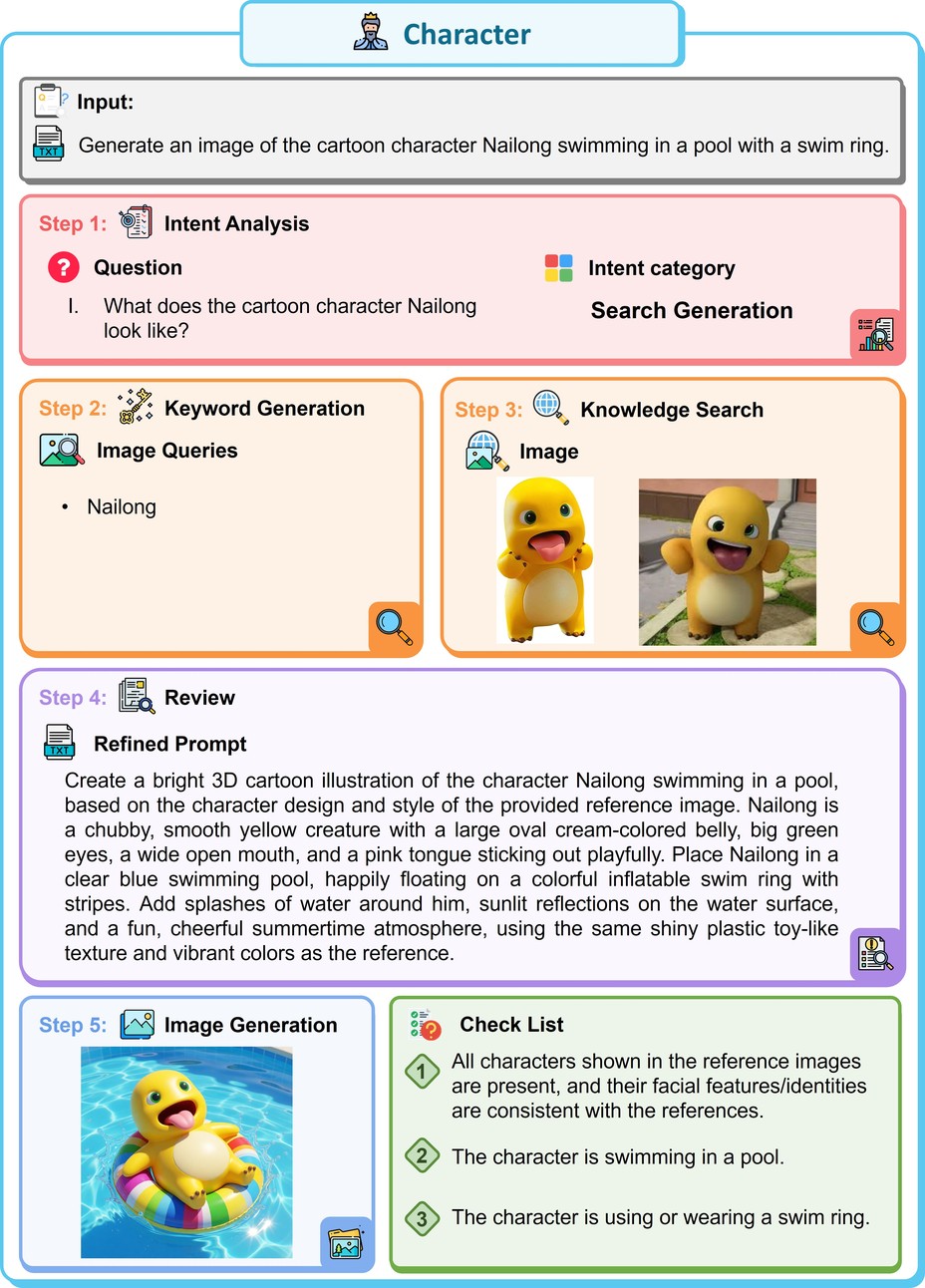}
    \caption{A generation process of Mind-Brush in Character task of Mind-Bench.}
    \label{fig:sup_Character1}
\end{figure*}

\begin{figure*}[t!]
    \centering
    \includegraphics[width=0.75\linewidth]{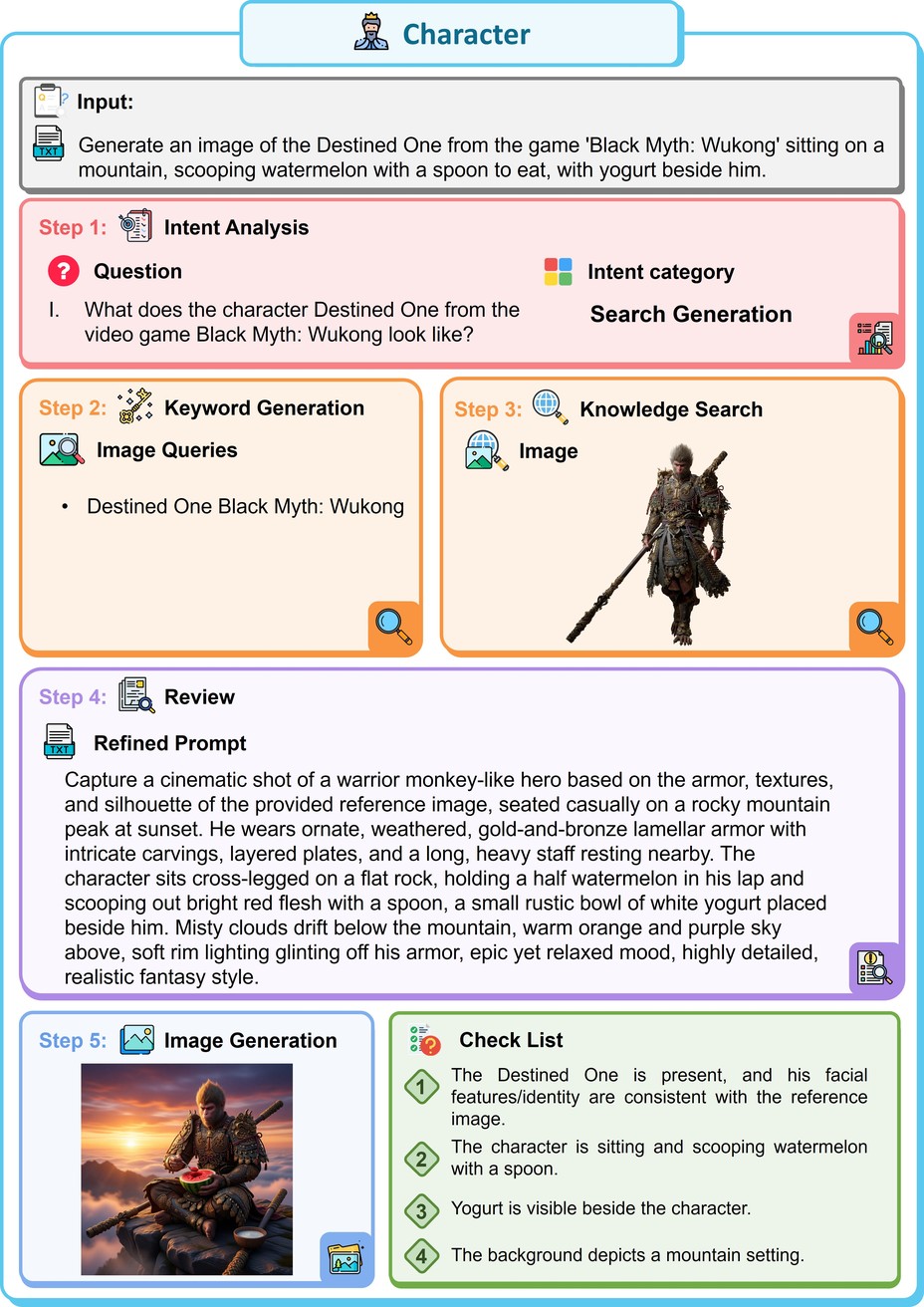}
    \caption{A generation process of Mind-Brush in Character task of Mind-Bench.}
    \label{fig:sup_Character2}
\end{figure*}

\begin{figure*}[t!]
    \centering
    \includegraphics[width=0.75\linewidth]{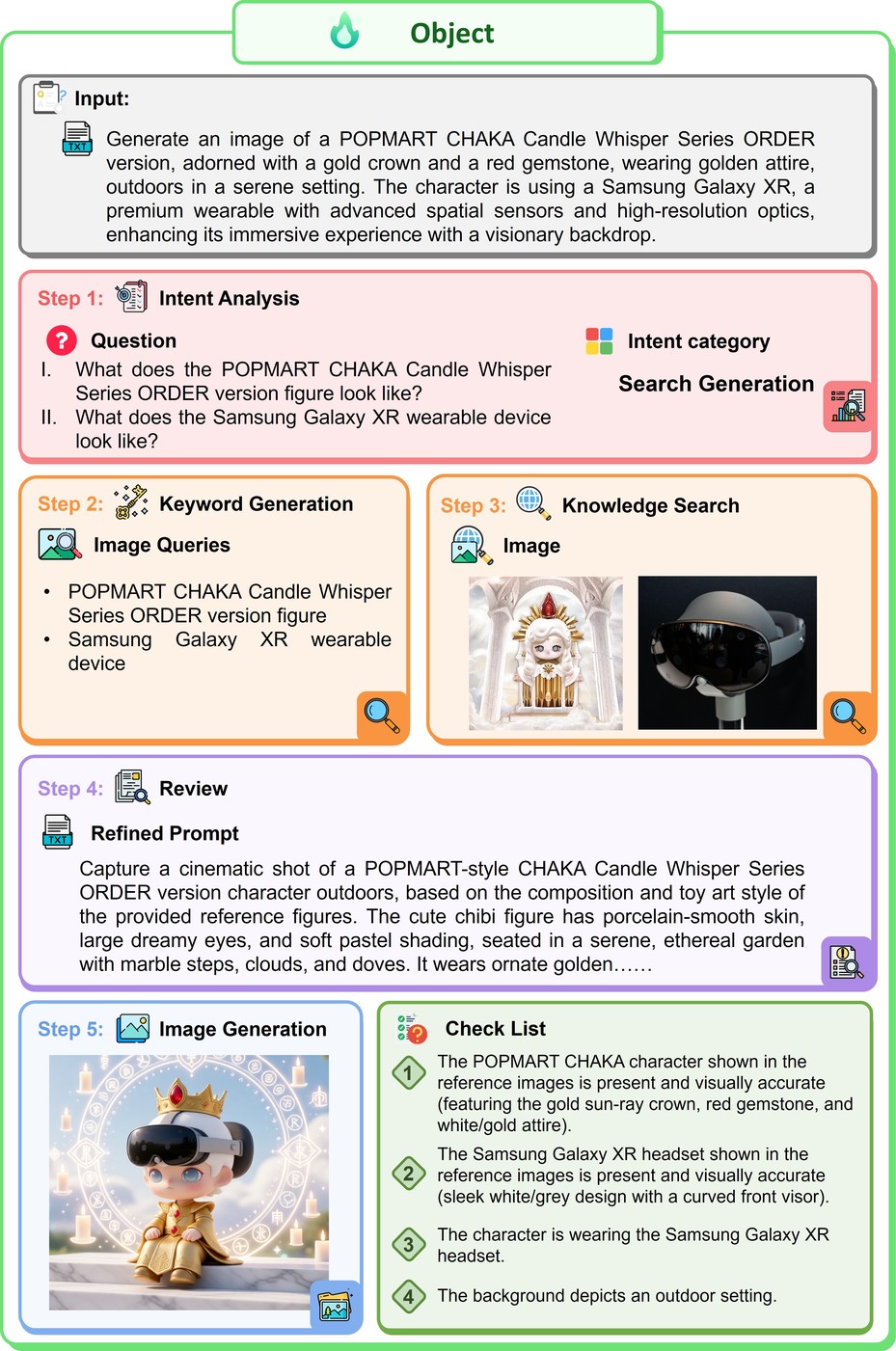}
    \caption{A generation process of Mind-Brush in Object task of Mind-Bench.}
    \label{fig:sup_object1}
\end{figure*}

\begin{figure*}[t!]
    \centering
    \includegraphics[width=0.75\linewidth]{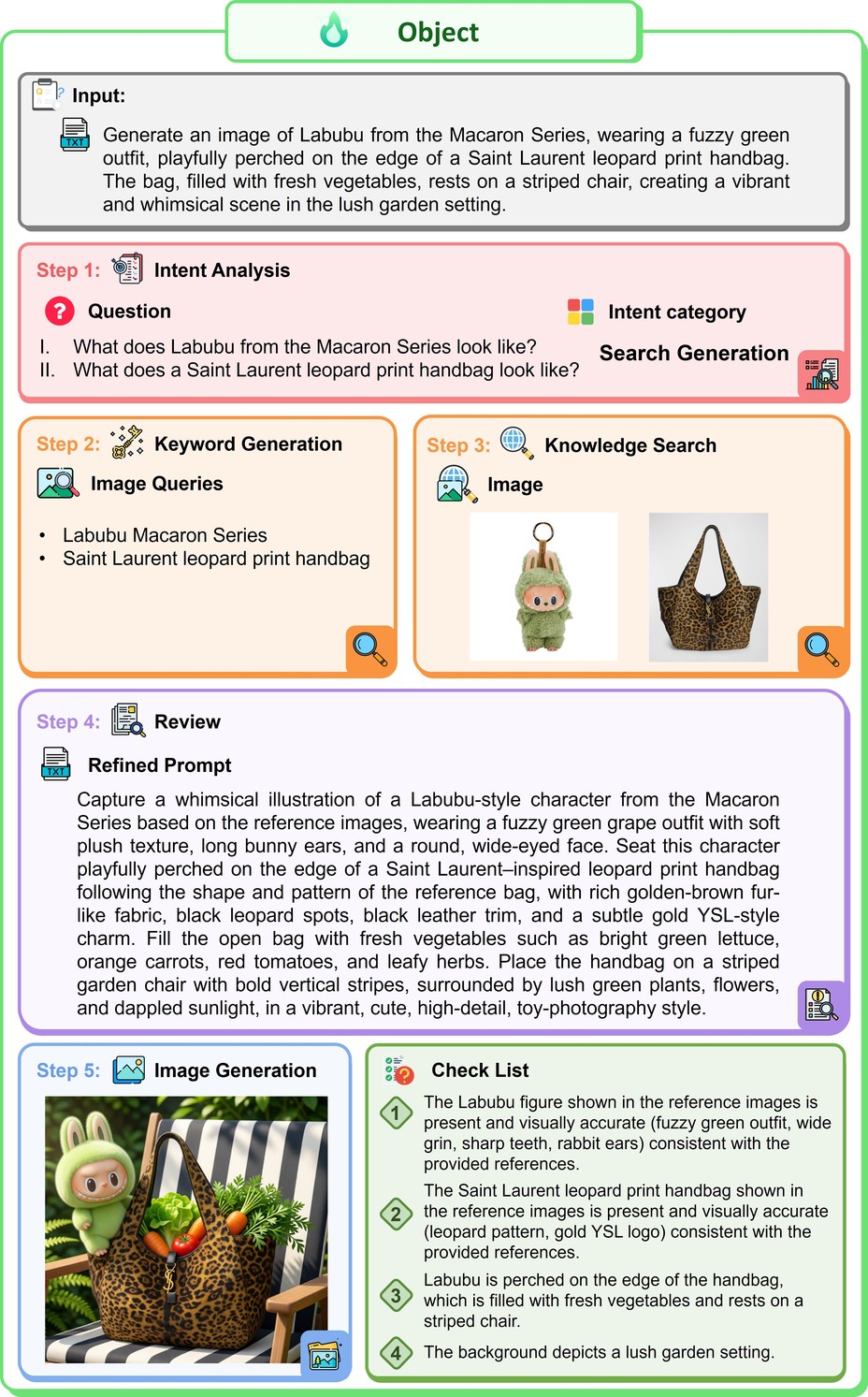}
    \caption{A generation process of Mind-Brush in Object task of Mind-Bench.}
    \label{fig:sup_object2}
\end{figure*}

\begin{figure*}[t!]
    \centering
    \includegraphics[width=0.75\linewidth]{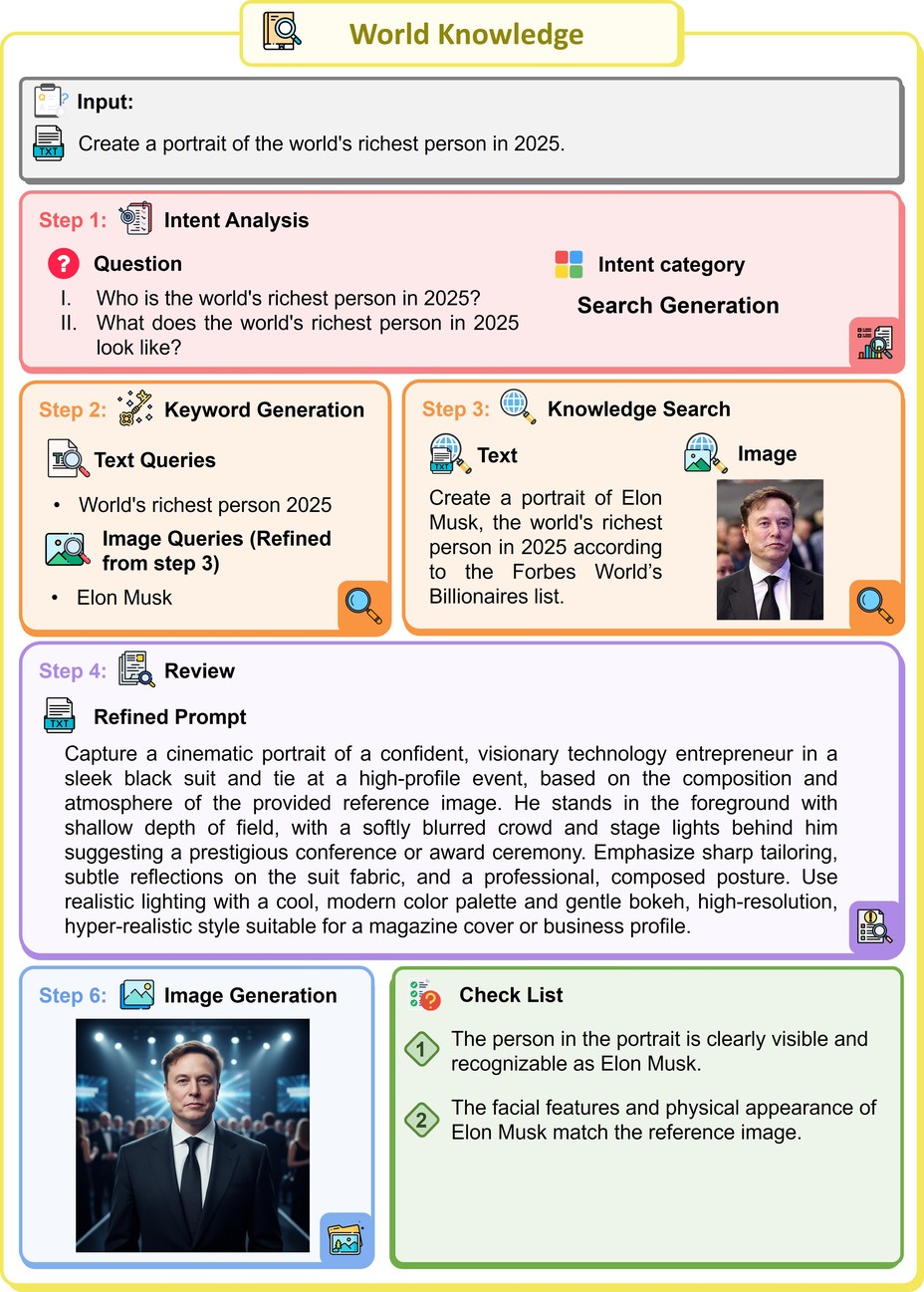}
    \caption{A generation process of Mind-Brush in World Knowledge task of Mind-Bench.}
    \label{fig:sup_World1}
\end{figure*}

\begin{figure*}[t!]
    \centering
    \includegraphics[width=0.75\linewidth]{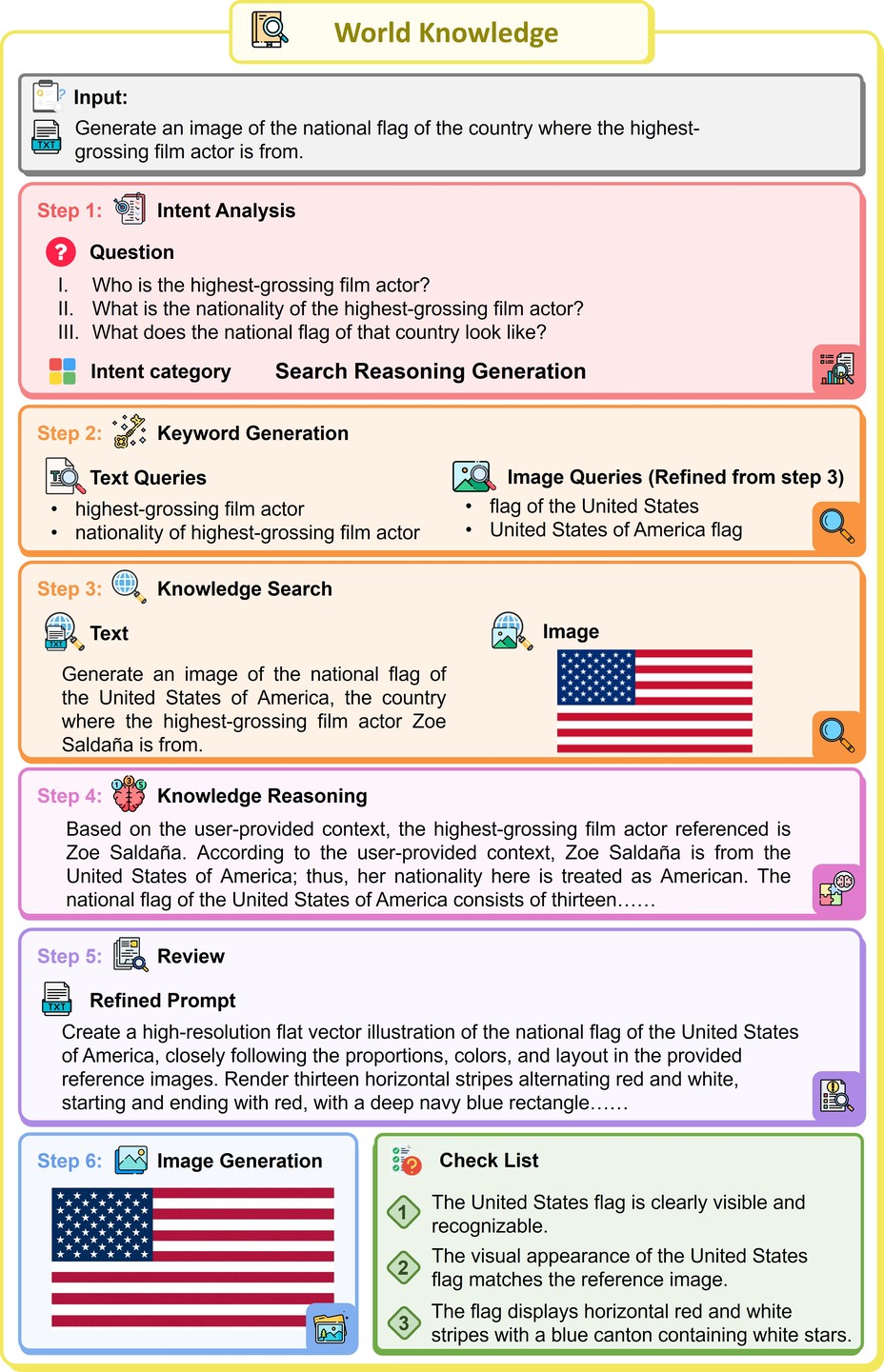}
    \caption{A generation process of Mind-Brush in World Knowledge task of Mind-Bench.}
    \label{fig:sup_World2}
\end{figure*}

\begin{figure*}[t!]
    \centering
    \includegraphics[width=0.75\linewidth]{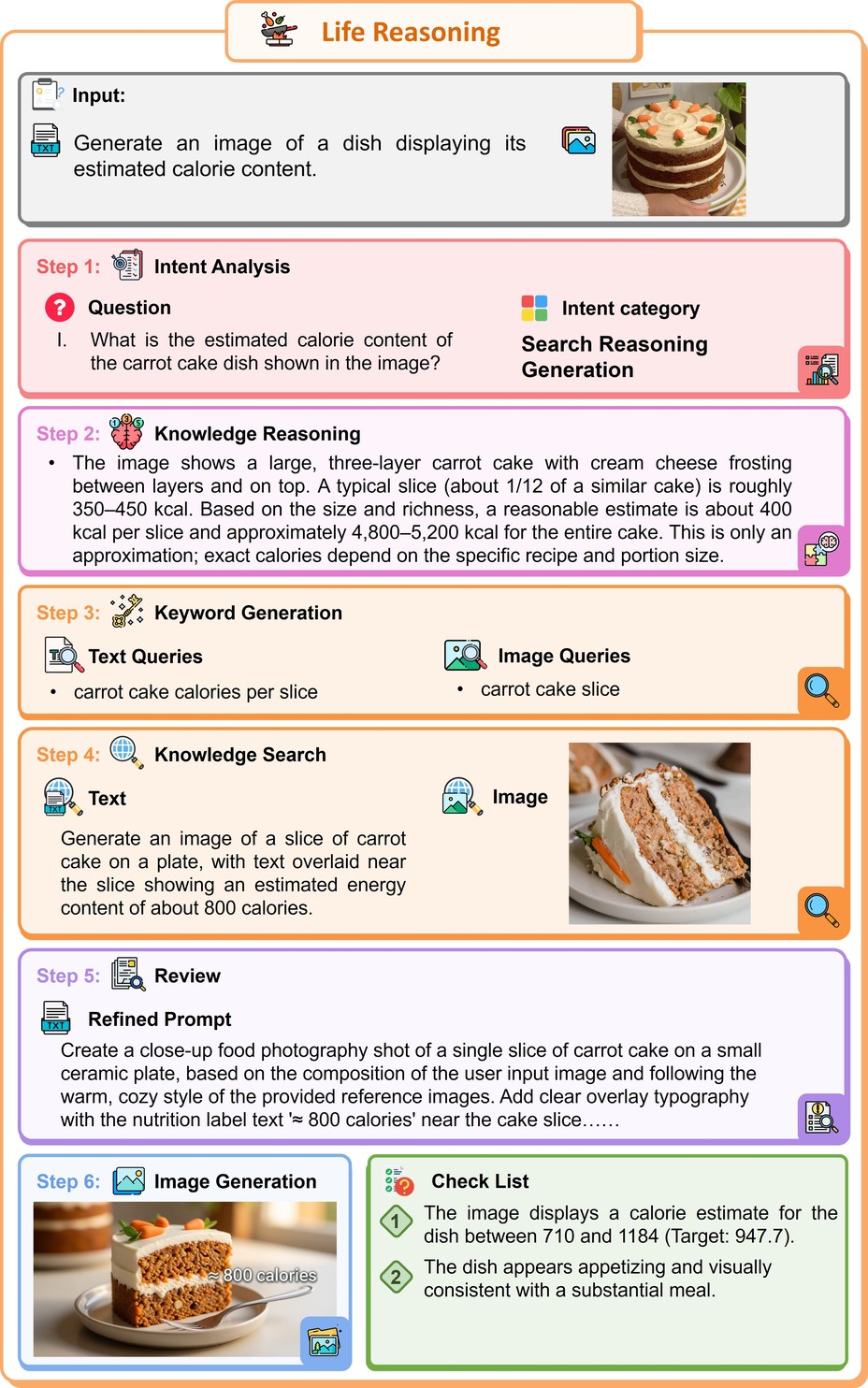}
    \caption{A generation process of Mind-Brush in Life Reasoning task of Mind-Bench.}
    \label{fig:sup_life1}
\end{figure*}

\begin{figure*}[t!]
    \centering
    \includegraphics[width=0.75\linewidth]{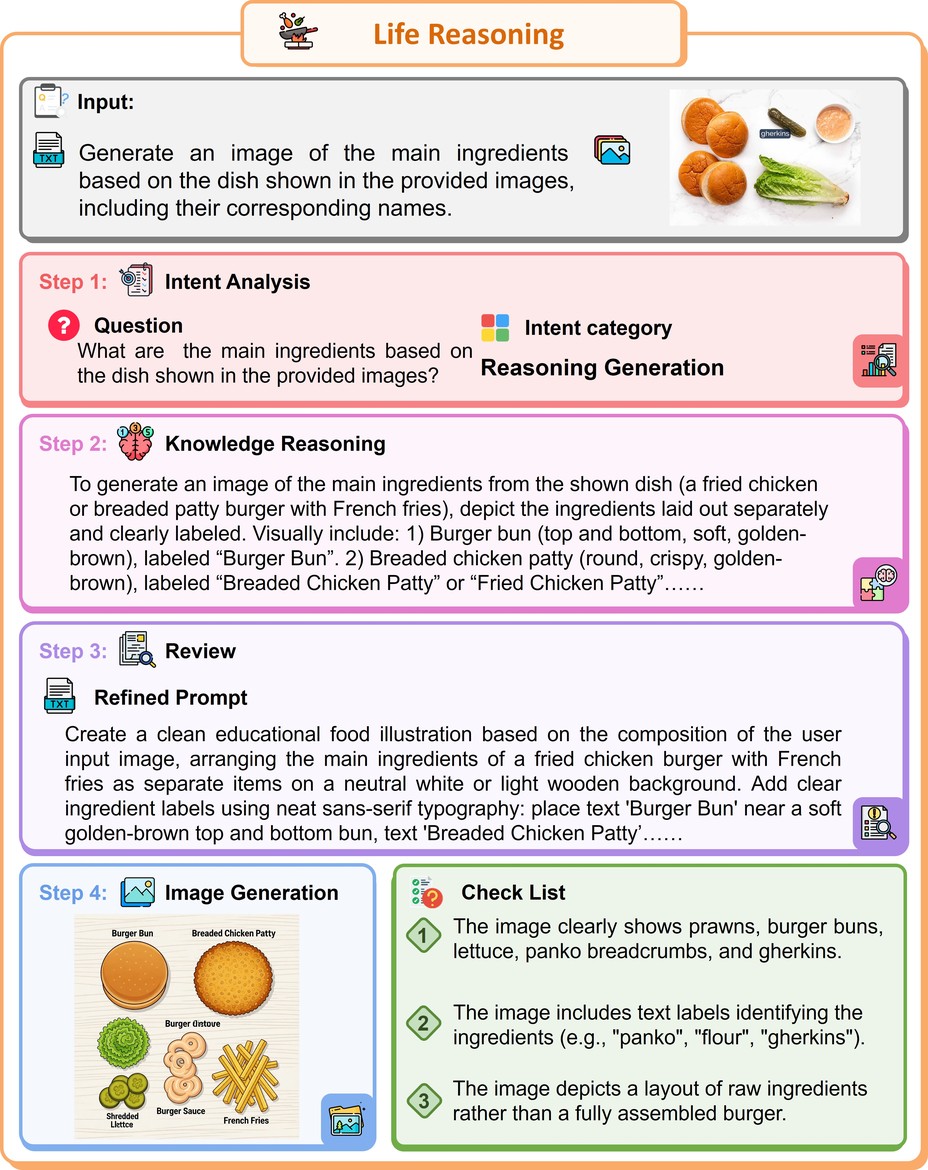}
    \caption{A generation process of Mind-Brush in Life Reasoning task of Mind-Bench.}
    \label{fig:sup_life2}
\end{figure*}

\begin{figure*}[t!]
    \centering
    \includegraphics[width=0.75\linewidth]{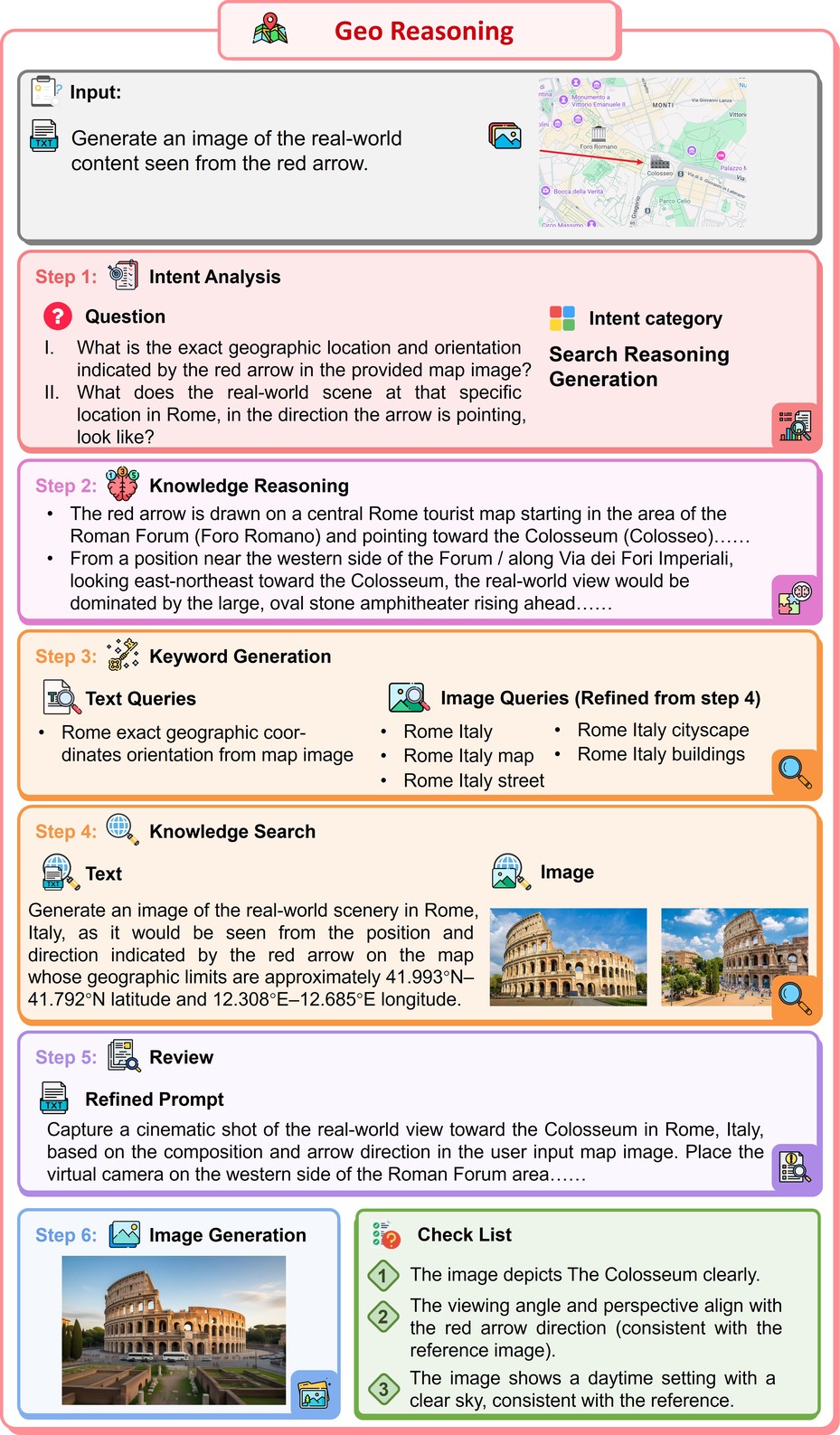}
    \caption{A generation process of Mind-Brush in Geo Reasoning task of Mind-Bench.}
    \label{fig:sup_Geo1}
\end{figure*}

\begin{figure*}[t!]
    \centering
    \includegraphics[width=0.72\linewidth]{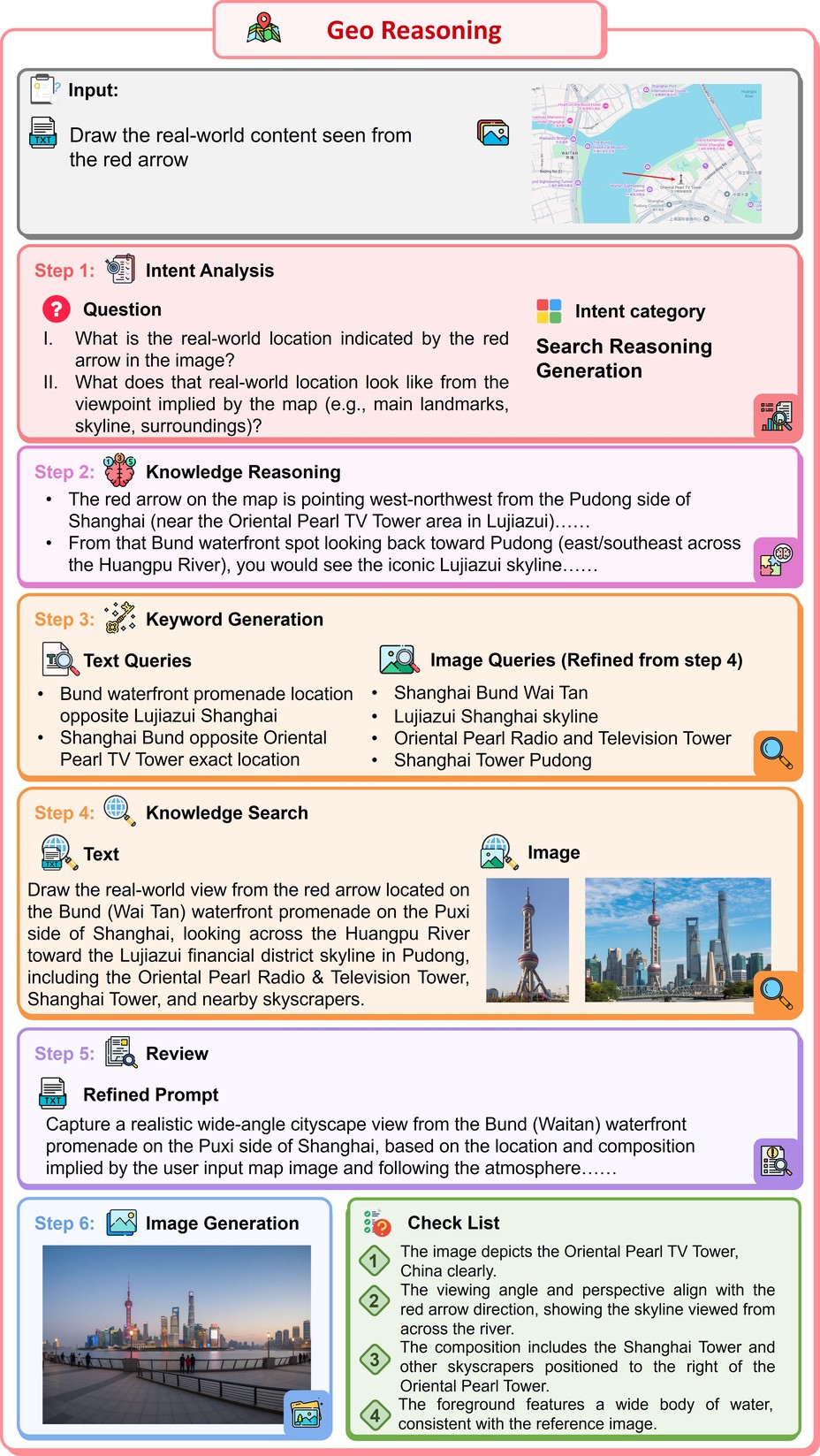}
    \caption{A generation process of Mind-Brush in Geo Reasoning task of Mind-Bench.}
    \label{fig:sup_Geo2}
\end{figure*}

\begin{figure*}[t!]
    \centering
    \includegraphics[width=0.75\linewidth]{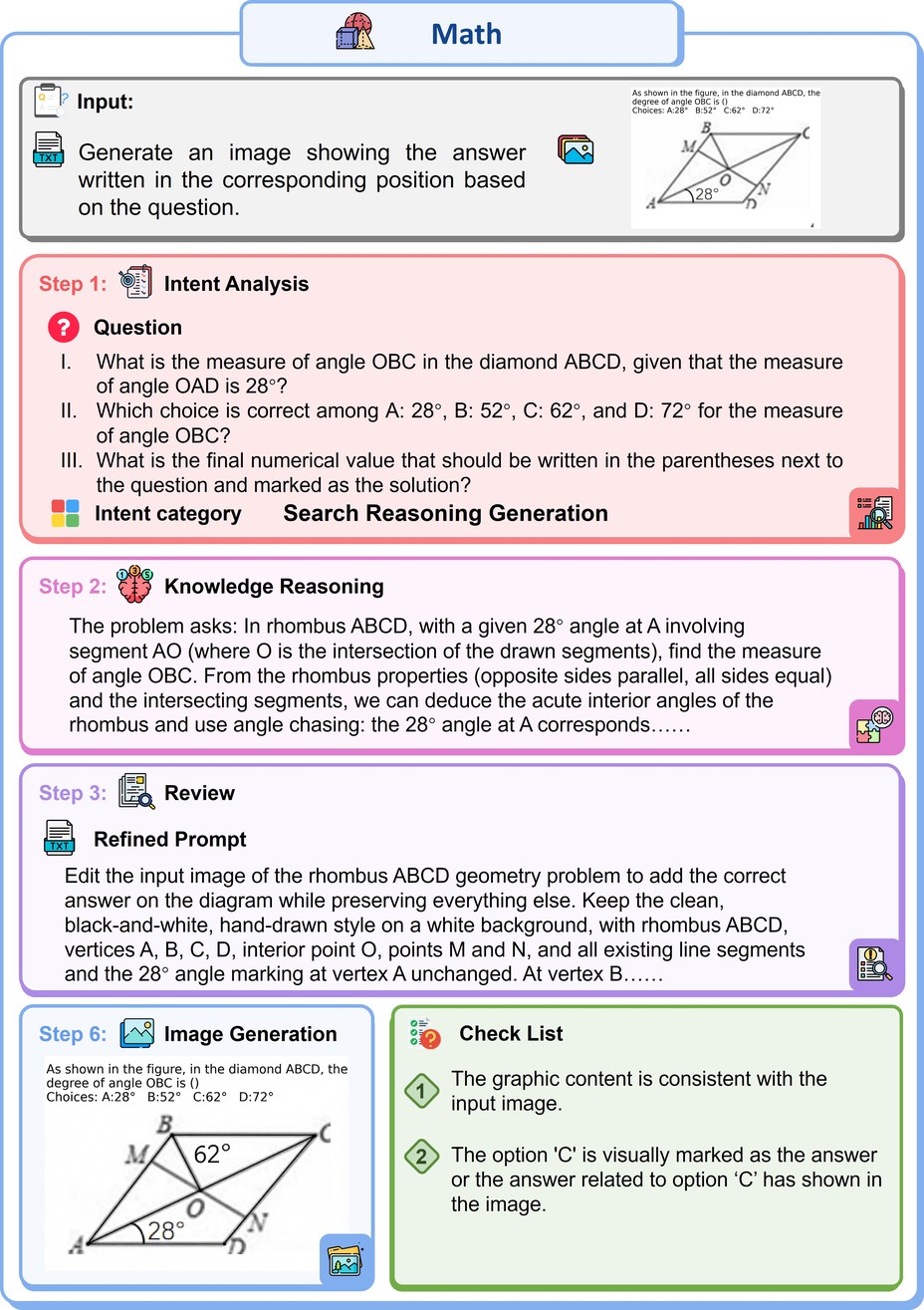}
    \caption{A generation process of Mind-Brush in Geo Math task of Mind-Bench.}
    \label{fig:sup_math1}
\end{figure*}

\begin{figure*}[t!]
    \centering
    \includegraphics[width=0.75\linewidth]{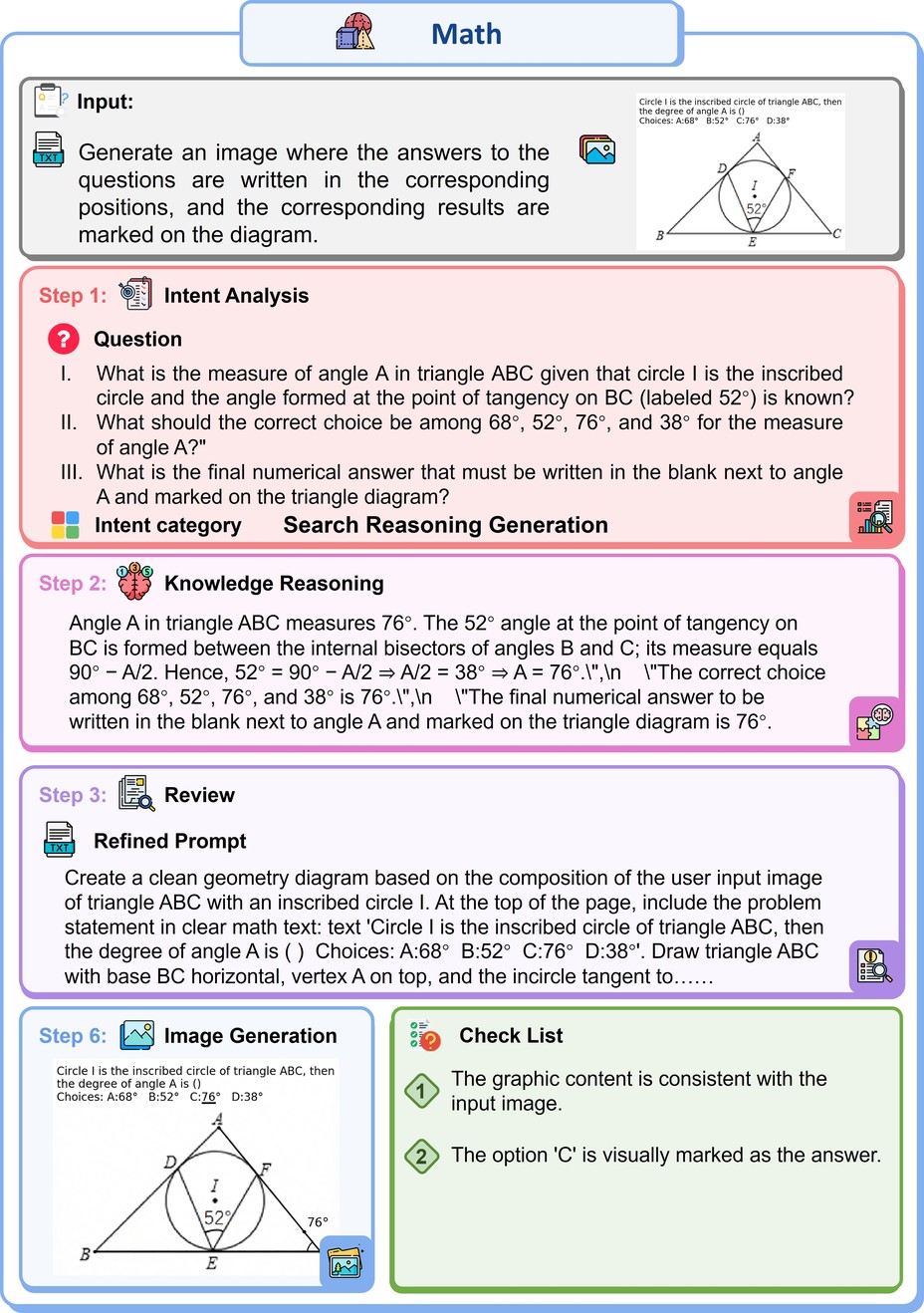}
    \caption{A generation process of Mind-Brush in Math task of Mind-Bench.}
    \label{fig:sup_math2}
\end{figure*}

\begin{figure*}[t!]
    \centering
    \includegraphics[width=0.75\linewidth]{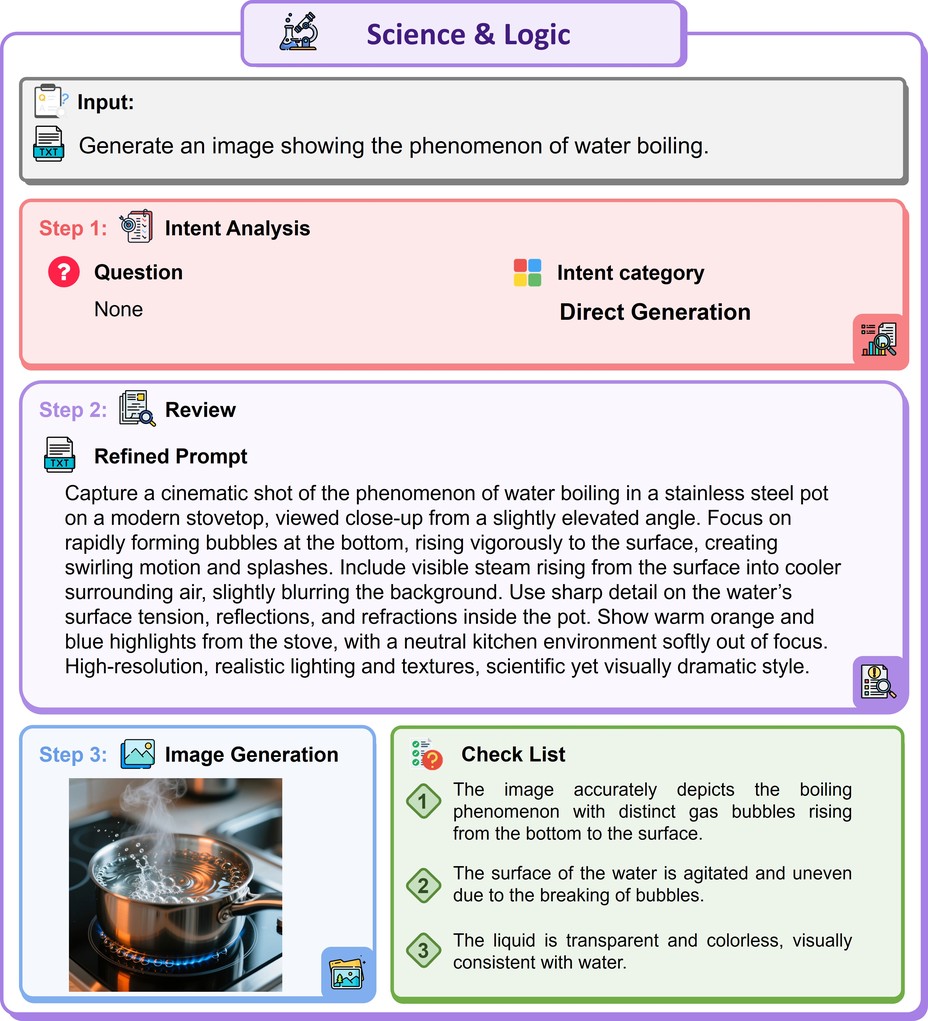}
    \caption{A generation process of Mind-Brush in Science \& logic task of Mind-Bench.}
    \label{fig:sup_science1}
\end{figure*}

\begin{figure*}[t!]
    \centering
    \includegraphics[width=0.75\linewidth]{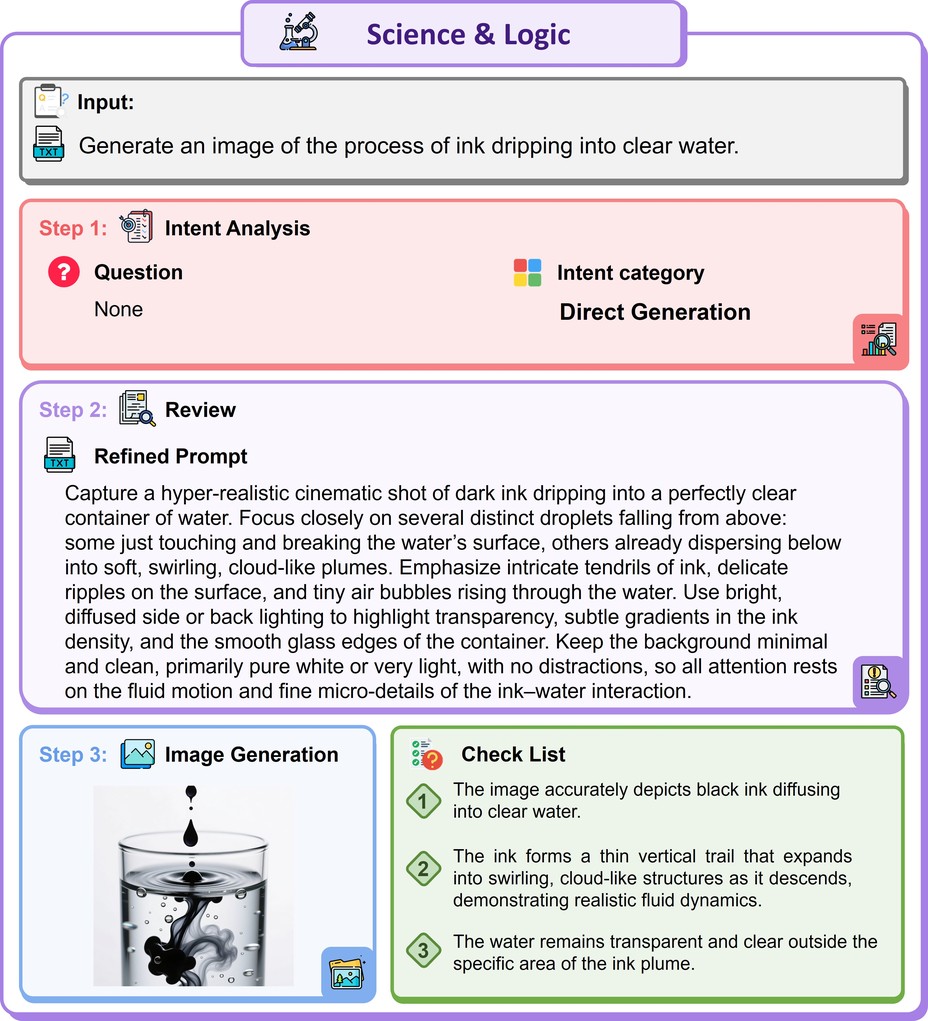}
    \caption{A generation process of Mind-Brush in Science \& logic task of Mind-Bench.}
    \label{fig:sup_science2}
\end{figure*}

\begin{figure*}[t!]
    \centering
    \includegraphics[width=0.75\linewidth]{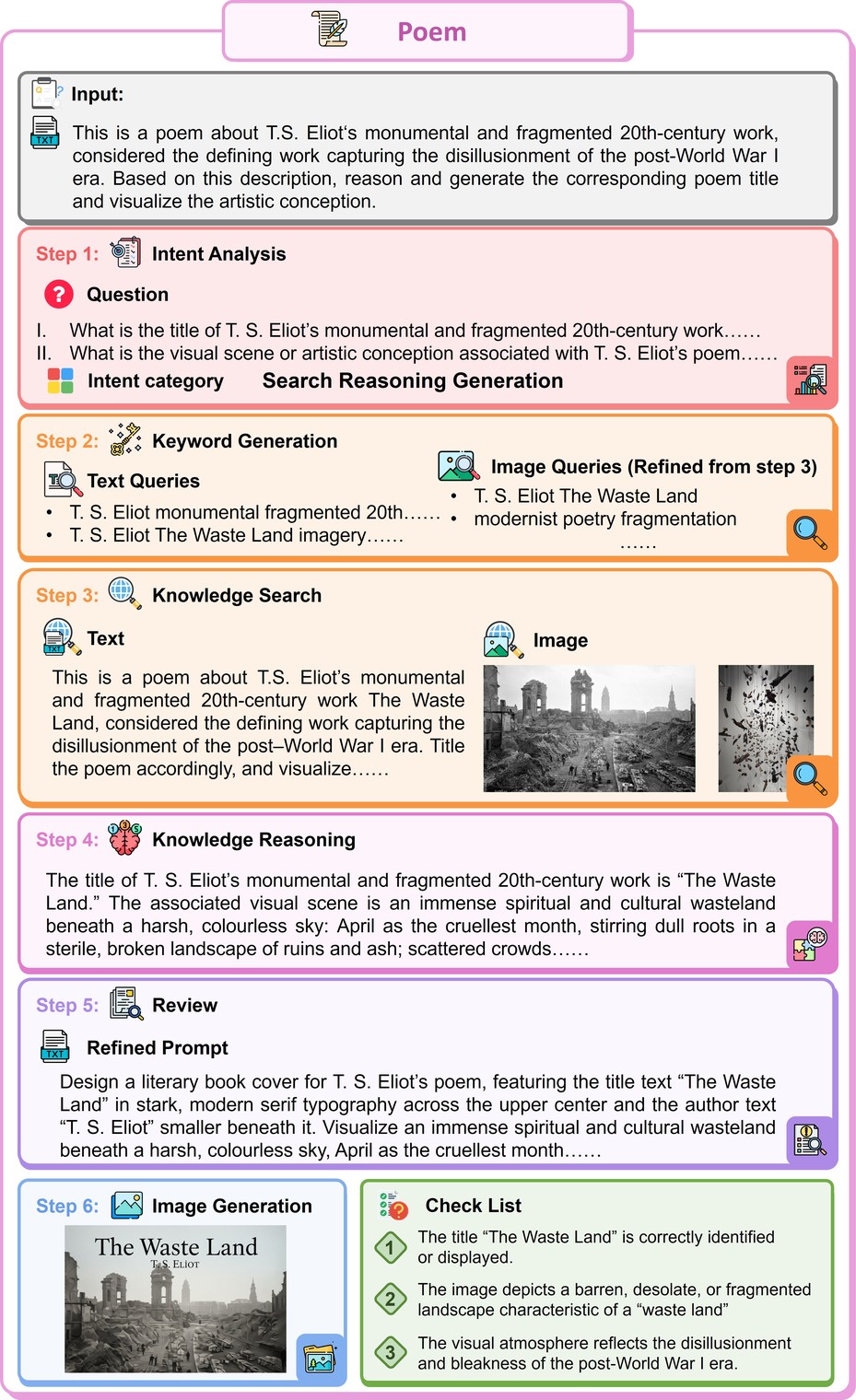}
    \caption{A generation process of Mind-Brush in Poem task of Mind-Bench.}
    \label{fig:sup_Poem1}
\end{figure*}

\end{document}